\pdfoutput=1

\documentclass[11pt]{article}

\usepackage[]{acl}

\usepackage{times}
\usepackage{latexsym}
\usepackage{enumitem}
\usepackage{adjustbox}  
\usepackage{booktabs}
\usepackage{subcaption}
\usepackage{graphicx}
\usepackage{multirow}
\usepackage{color, colortbl}
\usepackage{graphics}
\usepackage{amsmath}
\usepackage{lipsum}
\usepackage{algorithm}
\usepackage{algpseudocode}
\usepackage{comment}
\usepackage{makecell}
\usepackage{bm}
\usepackage{amssymb}
\usepackage{float}
\usepackage{booktabs}
\usepackage{tabularx}
\usepackage{ragged2e}
\newcolumntype{Y}{>{\RaggedRight\arraybackslash}X}
\usepackage[T1]{fontenc}

\usepackage[utf8]{inputenc}

\usepackage{microtype}

\usepackage[htt]{hyphenat}

\usepackage{listings}
\usepackage[most]{tcolorbox}

\title{Effects of Collaboration on the Performance of Interactive Theme Discovery Systems}

\author{
   Alvin Po-Chun Chen$^*$\;\;
   Rohan Das$^*$\;\;
   Dananjay Srinivas$^*$ \\
   \textbf{Alexandra Barry} \;\;
   \textbf{Maksim Seniw} \;\;
   \textbf{Maria Leonor Pacheco} \\
   University of Colorado Boulder \\
   \texttt{\{alvin.chen, rohan.das, dananjay.srinivas\}@colorado.edu} 
}

\begin{document}
\maketitle

\footnotetext[1]{Equal Contribution}

\begin{abstract}
NLP-assisted solutions to support qualitative data analysis have gained considerable traction. However, no unified evaluation framework exists which can account for the many different settings in which qualitative researchers may employ them. In this paper, we propose a framework to evaluate the way collaboration settings may produce different research outcomes across a variety of interactive systems. Specifically, we study the impact of synchronous vs. asynchronous collaboration using three different NLP-assisted qualitative research tools and present a comprehensive analysis of the differences in the consistency, cohesiveness, and correctness of their outcomes. 
\end{abstract}

\section{Introduction}\label{sec:intro} 

Making sense of large textual datasets is a common challenge across academic disciplines and is traditionally addressed through qualitative methods such as Thematic Analysis \citep{braun-2006} and Grounded Theory \citep{glaser1968discovery}. These approaches rely on manual \textit{inductive coding}, in which researchers identify abstract themes by closely reading the data. However, as datasets grow in size, manual coding becomes impractical, motivating the use of Natural Language Processing (NLP) techniques to automate parts of the analysis process~\cite{doi:10.1146/annurev-polisci-090216-023229,IJoC10675}.

In recent years, a range of NLP-based systems have been developed to support qualitative research. These systems assist researchers by uncovering latent semantic structures through topic modeling \cite{smith-2018-loop,fang-2023-user}, clustering documents and propagating limited human annotations across datasets \cite{pacheco-2023-interactive,chew2023llmassistedcontentanalysisusing}, or offering real-time coding recommendations \cite{dai-etal-2023-llm, gao-2024-collabcoder}. To maintain researcher agency, such systems typically adopt human-in-the-loop (HitL) strategies that balance automation with manual interpretation.

\begin{figure}[t!]
    \centering
    \includegraphics[width=1\linewidth]{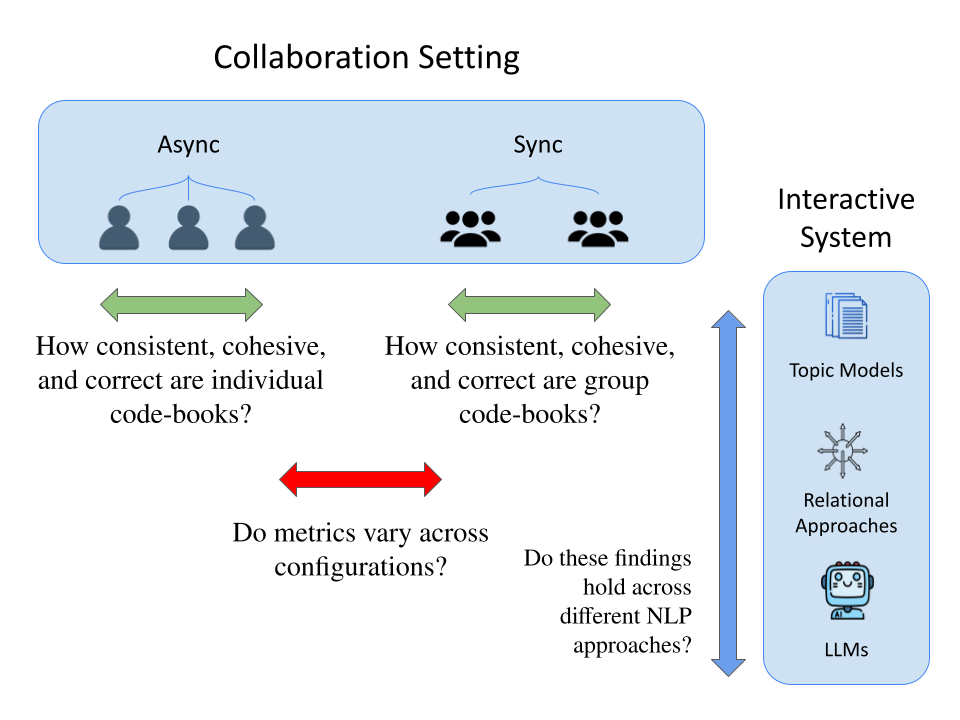}
    \caption{In this study, we measure the quality of coded themes using different interactive systems under different coding configurations.
    }
    \label{fig:overview}
\end{figure}

Previous work typically evaluates HiTL qualitative analysis tools in isolation, focusing on specific technical strengths and weaknesses. For example, by comparing topic coherence with and without human input \cite{fang-2023-user} or contrasting machine-assisted and manual code-book generation \cite{dai-etal-2023-llm}. However, qualitative analysis in practice is often collaborative, with teams of researchers jointly coding and interpreting data~\cite{cite-key}, and supported by tools built on diverse methodologies that are applied to datasets with vastly differing characteristics~\cite{Baden02012022}. By abstracting away collaboration settings, methodological variation, and dataset diversity, existing evaluations risk misrepresenting how HitL systems operate in real-world settings and how broadly their findings apply. In this work, we seek to answer the following questions: (1) Does the collaboration setting measurably affect the quality of resulting code-books? (2) Do these findings hold across different NLP approaches? (3) How do dataset characteristics influence the outcomes of NLP-assisted inductive coding tools?

We focus on two common but contrasting collaboration settings: asynchronous coding, where individuals code independently before consolidating results, and synchronous coding, where teams identify themes through live discussion. These settings offer complementary strengths, with asynchronous coding supporting flexibility across time and location, and synchronous coding facilitating shared understanding and efficient coordination. To compare outcomes across these settings, we introduce an evaluation framework that measures consistency between synchronous and asynchronous coding, as well as the cohesiveness and correctness of themes produced within each setting.

We evaluate three NLP-assisted inductive coding tools built on distinct methodological foundations: a human-in-the-loop topic modeling system \cite{fang-2023-user}, a concept-driven thematic modeling approach \cite{pacheco-2023-interactive}, and an LLM-based system for evaluating and propagating code definitions \cite{chew2023llmassistedcontentanalysisusing}. To assess generalizability across data characteristics, we test all tools and collaboration settings on two markedly different datasets: a corpus of short social media posts and a collection of advertising texts. We further complement our quantitative analysis with a small-scale user study to capture coder experiences and derive design recommendations.  


Our contributions are twofold: (1) we demonstrate that collaboration setting affects the results of NLP-assisted inductive coding tools across diverse methodologies and datasets, and (2) we provide an evaluation strategy that captures multiple dimensions of analysis quality. Together, these findings aim to inform the design and evaluation of language technologies that better align with real-world qualitative research workflows. Our code and experimental data are publicly available\footnote[2]{\scriptsize{\url{https://github.com/blast-cu/interactive-systems}}}.
\section{Related Work}
The overarching goal of the systems we investigate is to partially automate the qualitative coding process, either by inducing topics in an interactive, semi-supervised manner \cite{fang-2023-user, smith-2018-loop}, by learning user-defined themes interactively \cite{pacheco-2023-interactive, gao-2023-coaicoder}, or by prompting LLMs with natural language definitions of the observed themes \cite{chew2023llmassistedcontentanalysisusing, dai-etal-2023-llm}. A separate but related line of work exemplified by \citet{gao-2024-collabcoder} uses LLMs to generate label recommendations as users perform the coding process. While this system is explicitly designed for asynchronous collaboration, the systems we study differ in their ability to annotate large portions of the dataset without extensive supervision.

Our research addresses a real-world use case for qualitative researchers using HiTL systems and is informed by the Human-Computer Interaction (HCI) literature~\cite{jiang-2021-serendipity, feuston2021tools, chen-2018-ambiguity}. Prior HCI work shows that coders place particular emphasis on identifying and resolving ambiguity. In traditional settings, this is supported by an independent close reading of the data. However, in large-scale, NLP-assisted analysis, coders have limited visibility into where such ambiguities arise. Solutions have been proposed to either visualize codes \citep{drouhard-aeonium} or rank document disagreement \citep{zade-2018-disagree} regardless of dataset size. In contrast, our evaluation methods show different qualities of the resulting themes and document assignments by using signals from group overlaps, relationships in the semantic embedding space, and post-hoc evaluations. This highlights areas where the coders diverge both with each other and with the model, providing another perspective on the ambiguity question.

Previous evaluation methods introduced with HiTL systems for qualitative coding have generally been ad hoc, with experiments conducted in various group settings ~\cite{choo-utopian, hoque-convisit, smith-2018-loop}, on individual participants ~\cite{rietz-cody}, and through platforms such as MTurk ~\cite{zade-2018-disagree}. Our contribution provides a standardized framework for performing experiments regardless of the collaboration modality, using a suite of metrics for evaluating consistency, cohesiveness, and correctness in experimental results.
\section{Interactive Systems}\label{sec:systems} 
We identify three categories of NLP techniques used in interactive systems for qualitative coding with large datasets: topic models, relational approaches, and LLMs. Other techniques exist, but we focus on these three for their ability to help code large datasets. To maximize coverage across systems, we select a representative system from each category to use in our experiments. In this section, we briefly describe the unique aspects of each category and introduce the selected system.

\subsection{Topic Models}
Tools in this category use some variation of topic modeling to find emerging themes and facilitate document assignment. These systems benefit from the relative speed of the topic model, which allow users to quickly visualize and explore the dataset. Early exploration incorporated visualizations to help users adjust parameters~\citep{Chuang2013DocumentEW}, while later works implemented refinement operations that allow users to directly edit topic words and remove documents~\citep{smith-2018-loop}. However, topic modeling systems are limited by their lack of malleability and predictability. Refinement operations mostly edit topic words, which can have limited impact in the final results since users cannot directly reassign documents. 

We select the HitL query-driven topic model (QDTM) system introduced by \citet{fang-2023-user}. This topic model is initialized by providing input queries (words that represent concepts of interest for the user) which the model uses to generate the initial topics. Users begin by iterating through each topic and naming them based on identified themes. They then use a set of \textit{refinement operations} to edit the topic model. These include: merging and splitting topics based on topic words, adding, removing, or reordering topic words, and removing documents from topics. The next iteration of the model is only produced when the users choose to apply refinements. Each iteration is saved, allowing users to return to prior iterations to test different operations. Once satisfied with the state of the topic model, the user downloads the document distribution for that iteration.

To ensure comparable results, we use the same starting distribution of 13 topics for all our experiments using the same hyperparameters as \citet{fang-etal-2021-query} ($\alpha=1.0$, $\beta=0.5$, $\gamma=1.5$) and without any input queries. The same initial topic model is provided for all experiments. 

\subsection{Relational Approaches}\label{sec:relational}
Relational approaches combine vector semantics and structured inference to model relationships between high-level concepts. Instead of treating themes as distributions over words (as topic models do), these frameworks define themes as distributions over generalized concepts. This reflects the inductive coding process, where researchers identify patterns and concepts that are then synthesized into more abstract themes. However, their computational complexity grows with the number of dependencies considered, which hinders their ability to quickly adapt during coding sessions. Further, they rely on users to define informative concepts, making them less suited for inexperienced researchers.  

We select the relational system introduced by \citet{pacheco-2023-interactive}, which uses a two-stage relational framework. In the first stage, the system automatically partitions the dataset based on semantic similarity. We use SBERT~\citep{reimers-gurevych-2019-sentence} to embed each document and partition the dataset using K-Means clustering~\citep{Jin2010}. The users explore each partition to identify themes, assign "good" and "bad" example documents for each theme, and input or correct supporting concepts for each example. In the second stage, the system uses the provided examples and concept relations to map the remaining dataset, only leaving documents unmapped if no theme is a sufficiently good match. Assignments are produced by a structured inference procedure, formulated as an approximate integer linear program, that explicitly models dependencies between concepts and themes. The unmapped documents are repartitioned as in the first stage and users are prompted to review unmapped partitions again. The process is performed iteratively until all documents are mapped or until users are satisfied. 

In our experiments, researchers are provided with initial clusters to identify themes. We generate $K=10$ clusters to form the initial partitions, which are kept consistent across experiments. Within each partition, users name the themes they observe and select positive and negative examples for each. These examples are embedded and used to score semantic similarity with unlabeled documents. In parallel, users annotate each of these examples with its supporting concepts, which are then used to learn concept predictions over the corpus. The framework assumes a soft rule of the form \textit{concept} $\Rightarrow$ \textit{theme} for every concept-theme pair, with rule weights learned from the annotated examples. For instance, when coding the COVID-19 vaccine debate, the rule \textit{anti-vax} $\Rightarrow$ \textit{Natural Immunity is Effective} would receive a high weight if most examples of that theme were annotated with an anti-vax stance. The rules and their learned weights are then combined into a structured inference task that predicts label assignments jointly across the corpus. By drawing on both distributional similarity and learned concept-theme dependencies, the model extrapolates efficiently from a small set of manually labeled examples. Table~\ref{tab:ops} in Appendix \ref{app:drail_details} details all operations available to users during coding.

\subsection{Large Language Models} 
LLMs are ideal candidates for interactive systems, especially for tasks such as qualitative coding where the model can be prompted to produce themes or explanations without ad-hoc training \citep{llms0shotreasoners}. They have been used for theme recommendation \citep{gao-2023-coaicoder}, for code conflict resolution \citep{gao-2024-collabcoder}, and for automated document assignment \citep{deductivecodingwithLLM}. However, the flexibility of LLM outputs also leads to hallucinations, which are only partially addressed by prompt engineering. Models further suffer from biases in training which are difficult to identify and can impact their ability to produce quality labels or recommendations \citep{chen-2018-ambiguity}. Additionally, their massive size is prohibitive when working with large datasets due to the high cost of inference.

We select the framework introduced by \citet{chew2023llmassistedcontentanalysisusing}. Here, the human coder first manually codes a representative subset of the data and drafts definitions for each code. The LLM is then prompted to label the data sample with the provided definitions. Agreement is calculated between human and model annotations using GWET's $AC_1$ \citep{gwet2008computing}. The prompt is then tweaked iteratively to achieve a satisfactory level of agreement, and the best-performing version is used to prompt the model to code the rest of the dataset. 

For our experiments, we select Llama 3.2 3B-Instruct as the base LLM for automated labeling. We use the same starting partitions as in the relational approach (Sec. \ref{sec:relational}), and themes are instantiated following the same process, except that users provide richer textual definitions for each theme rather than positive and negative examples and concept annotations. Additional details about our experimental settings can be found in Appendix~\ref{app:llm_details}. 

\section{Study Design}

To study the effects of collaboration settings on the performance of the three selected systems, we design a protocol that can be used for both synchronous and asynchronous settings. For each system, we conduct three asynchronous experiments with one coder each and two synchronous experiments with three coders each for a total of 30 experiments (15 per dataset). Evaluation metrics are calculated by comparing the resulting code-books within each experimental setting (e.g. the two code-books independently created by the two synchronous groups using the topic model). The rest of this section details the datasets, participant demographics, and experimental protocols.


\subsection{Datasets} We perform our experiments using two distinct datasets. The first consists of 85,799 tweets about COVID-19 vaccines posted by users located in the US, uniformly distributed between Jan.-Oct.~2021 \citep{pacheco-etal-2022-holistic}. The corpus also contains labels for vaccination stance (e.g. pro-vax, anti-vax) and morality frames (e.g. fairness/cheating and their actor/targets)~\cite{roy-etal-2021-identifying}, which are used as auxiliary concepts for the relational model. The second dataset consists of 5,471 climate related English language ads from the Facebook Ad Library~\cite{islam-2023-climate}. The other dataset focuses on Facebook ads shown in the US between Jan.~2021-Jan.~2022, and contain labels for climate change stance.  The two corpora differ substantially in length, complexity, and semantic structure: COVID tweets are short and semantically homogeneous, while climate ads are longer, more linguistically complex, and more formulaic. We characterize these differences quantitatively in Appendix~\ref{app:dataset} and discuss their implications in our evaluation (Section~\ref{sec:dataset_differences}). 

\subsection{Participants} We recruited a group of 33 researchers in NLP and Computational Social Science, 9 female and 24 male, between the ages of 20 and 45. This group included professors at different levels of seniority, postdoctoral researchers, and graduate and undergraduate students from two different universities. This group covers the range of researchers likely to use interactive coding systems. All participants were either well-versed in qualitative data analysis, or were explicitly trained by senior researchers to perform the task. Due to the large number of experiments in our study, some participants took part in multiple experiments. These participants always performed the asynchronous experiment first to prevent external influence and took part in at most one synchronous experiment.

\begin{table*}[ht]
\centering
\resizebox{\textwidth}{!}{%
\begin{tabular}{@{}ll|lll|lll|lll@{}}
\toprule
      & & \multicolumn{3}{c|}{Topic Model} & \multicolumn{3}{c|}{Relational} & \multicolumn{3}{c}{LLM-Based} \\
      & & Jaccard & Centroid & Group Avg. & Jaccard & Centroid & Group Avg. & Jaccard & Centroid  & Group Avg. \\ 
      \midrule
      \multirow{2}*{COVID} & Sync  & $\bf0.56 (0.23)$ & $\bf0.98 (0.05)^{**}$ & $\bf0.52 (0.10)$ & $\bf0.36 (0.19)$ 
      & $\bf0.98 (0.01)^*$ & $\bf0.52 (0.07)^{*}$ & $0.14 (0.08)$ & $0.98 (0.03)$ 
      & $0.44 (0.03)$ \\
      & Async & $0.30 (0.17)$ & $0.96 (0.05)^{**}$ & $0.51 (0.09)$ & $0.30 (0.22)$ 
      & $0.94 (0.07)^{*}$ & $0.44 (0.10)^{*}$ & $\bf0.17 (0.11)$ & $0.98 (0.02)$ 
      & $\bf0.45 (0.03)$ \\ 
      \midrule
      \multirow{2}*{Climate} & Sync  & $0.37(0.31)^{*}$ & $0.89(0.14)$ & $0.43(0.14)$ & $0.27(0.17)$ & $\bf0.95(0.05)^{**}$ & $\bf0.43(0.08)$ & $0.09(0.07)^*$ 
      & $\mathbf{0.95(0.04)}$ & $\mathbf{0.37(0.07)}$ \\
      & Async & $\mathbf{0.58(0.29)^{*}}$ & $\mathbf{0.93(0.12)}$ & $\mathbf{0.46(0.16)}$ 
      & $\bf0.33(0.22)$ & $0.94(0.09)^{**}$ & $0.42(0.08)$ & $\mathbf{0.13(0.07)^*}$ 
      & $0.94(0.06)$ & $0.36(0.06)$ \\
\bottomrule
\end{tabular}}
\caption{Avg. Consistency between Best Theme Matches \textit{across} Coder Groups (standard deviation in brackets). \textbf{Bold} figures highlight higher consistency value between collaboration settings. $^*$Statistically significant using a two-sample unpaired t-test with $p<0.05$. $^{**}$ Near statistically significant with $p\approx0.05$.}
\label{tab:consistency}
\end{table*}

\subsection{Coding Protocol} At the start of each experiment, participants were provided with a demonstration of all the operations in their respective systems. Every system starts with an initial partition of the data, so participants were instructed to read the first 25 samples in each partition, and manually create/name any themes they identified before freely exploring the rest of the dataset and start performing operations to find more themes. 

In the topic model experiments, we suggested that participants merge and split topics based on their identified themes before making fine-grained refinements. They were then asked to refine the topic model based on their identified themes such that every topic corresponds to a unique theme. They kept re-running the model and making refinements until they were satisfied with the results, or until they failed to effect any meaningful changes. 

In the relational system experiments, participants were tasked with selecting example documents for each identified theme, as well as determining concept relations for them. Following \citet{pacheco-2023-interactive}, the supporting concepts considered were vaccination stances and morality frames (e.g., the identified theme ``natural immunity'' has an ``anti-vax'' stance, and is tied to the ``purity'' frame). Once participants were satisfied with their themes and selections, the system automatically coded the rest of the dataset. Unmapped examples were repartitioned and returned to the participants for a second (and last) round of coding. 

In the experiments for the LLM-based system, participants produced natural-language definitions for each identified theme and selected a set of good examples for them. We then prompted the LLM with different task-prompt templates to find the best prompt for each set of participant-generated definitions, which was then used to code the rest of the dataset. Details of the templates, as well as the human-model agreement for the best template can be found in Appendix \ref{app:llm_details}. 
\section{Evaluation}
\label{sec:eval}
We use both descriptive metrics and a user study to provide a comprehensive analysis on the differences when coding in synchronous and asynchronous settings. Our evaluation framework is comprised of three dimensions; \textbf{consistency}, \textbf{cohesiveness \& distinctiveness}, and \textbf{correctness}, each of which uses metrics that are well-established in the literature~\cite{NIPS2008_beed1360,10.5555/3540261.3540416,pacheco-2023-interactive}. The themes identified in each experiment are presented in Appendix \ref{app:theme-counts}. 

\begin{figure}[t]
    \centering
    \includegraphics[width=1\linewidth]{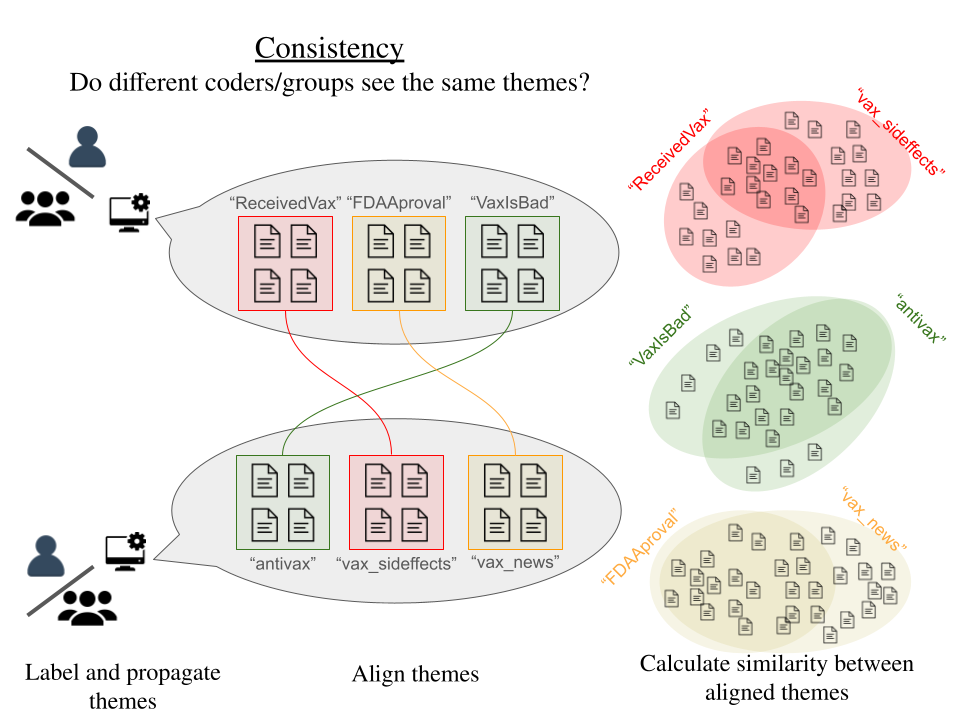}
    \caption{Two sets of coders use a particular HiTL system to find themes. Since the same theme can be named differently by different coders, we find the best match. 
    In this example, the coder 1's theme \texttt{``VaxIsBad''} has been matched with coder 2's theme \texttt{``antivax''}. 
    After aligning, we calculate the similarity between these two themes using methods like Jaccard Similarity or Centroid Distance.}
    \label{fig:consistency}
\end{figure}

\subsection{Consistency} Coders risk overgeneralizing or overlooking key themes, leading to unsystematic results~\citep{cornish-2014-sage-cqa}. We address this by measuring consistency, defined as the extent to which different coders elicit the same themes from the same texts (Fig.~\ref{fig:consistency}). In semi-automated coding, consistency is nontrivial to assess: similarly named themes may cover different documents, while differently named themes may overlap substantially. We therefore measure consistency based on document overlap between themes. Specifically, we compute the maximum Jaccard similarity between each theme and all themes produced by another coder, treating this maximum as the theme's best alignment. Consistency is then calculated as the average similarity across all aligned theme pairs. Example alignments can be found in Appendix \ref{app:jaccard_examples}.

To account for semantically similar themes with differing document assignments, we also measure semantic consistency using SBERT document embeddings~\cite{reimers-gurevych-2019-sentence}. We compute (1) centroid similarity, based on the cosine similarity between theme centroids, and (2) group average similarity, based on the average pairwise similarity across documents in two themes. As with Jaccard, we report maximum similarities per theme and average them for comparison across settings.

Table \ref{tab:consistency} reports average maximum Jaccard and embedding similarities across experiments. The high variance in Jaccard similarity across experiments underscores the importance of semantics-based metrics. For example, the topic model results on the COVID dataset show a high difference between collaboration settings: 0.56 average maximum Jaccard similarity in the synchronous experiment compared to 0.30 for the asynchronous one. However, the other two metrics show a much lower difference, suggesting a lower real difference in consistency.

For the COVID dataset, synchronous groups using the topic model and relational systems produced more consistent themes. The LLM-based system showed no statistically significant difference across collaboration settings and has less explanatory power. Results for the Climate dataset were less conclusive, likely due to greater semantic diversity and longer documents which may have encouraged synchronous coders to surface more varied themes through discussion; as shown in the next section, these themes are also more cohesive.

Notably, the LLM-based system afforded the fewest opportunities for user intervention during coding. Unlike the other systems, which supported operations such as splitting and merging topics or defining relations between concepts, the LLM system only allowed users to edit theme definitions. We hypothesize that richer intervention mechanisms better enable coders to leverage the deliberation inherent to synchronous collaboration.

\begin{table*}[ht]
\small
\centering
\resizebox{\textwidth}{!}{%
\begin{tabular}{@{}cll|ll|ll|ll@{}}
\toprule
    & & & \multicolumn{2}{c|}{Topic Model} & \multicolumn{2}{c|}{Relational}          & \multicolumn{2}{c}{LLM-based}                      \\
    & & & Intra-Theme & Inter-Theme & Intra-Theme & Inter-Theme          & Intra-Theme & Inter-Theme \\ 
    \midrule
    \multirow{4}*{COVID} & \multirow{2}*{All} & Sync & $0.52 (0.10)$ & $0.40(0.04)$            & $\bf0.51 (0.08)^{*}$ & $0.42 (0.05)^{*}$ & $\bf0.44 (0.06)$       & $0.40 (0.04)$ \\
    & & Async & $0.52 (0.10)$ & $0.40(0.04)$ & $0.45 (0.10)^{*}$         & $\bf0.34 (0.11)^{*}$ & $0.43 (0.05)$ & $\bf0.39 (0.04)$ \\
    & \multirow{2}*{Top 25\%} & Sync  & $0.56(0.11)$ & $0.39 (0.05)$     & $\bf0.70 (0.09)^{*}$ & $0.52 (0.07)^{*}$ & $0.63 (0.07)$          & $0.55 (0.05)$ \\
    & & Async & $0.56(0.11)$ & $0.39 (0.05)$ & $0.64 (0.09)^{*}$         & $\bf0.46 (0.13)^{*}$ & $0.63 (0.05)$ & $\bf0.54 (0.05)$ \\
    \midrule
    \multirow{4}*{Climate} & \multirow{2}*{All} & Sync & $\mathbf{0.57(0.23)^{*}}$ 
                           & $0.25(0.08)^*$ & $\bf0.44(0.10)^{*}$ & $0.30(0.07)^{*}$ 
                           & $\mathbf{0.39(0.11)^{*}}$       & $0.29(0.05)^{*}$ \\
    & & Async & $0.50(0.19)^{*}$ & $\mathbf{0.24(0.07)}^*$ & $0.43(0.09)^{*}$
    & $\bf0.29(0.07)^{*}$ & $0.38(0.08)^{*}$ & $\mathbf{0.28(0.05)}^{*}$ \\
    & \multirow{2}*{Top 25\%} & Sync  & $\mathbf{0.72(0.21)^{*}}$ & $0.26(0.09)^{*}$ 
    & $\bf0.66(0.11)^{*}$ & $0.39(0.10)^{*}$ & $\mathbf{0.63(0.11)^{*}}$ & $0.40(0.09)^{*}$ \\
    & & Async & $0.65(0.20)^{*}$ & $0.26(0.08)^{*}$ & $0.65(0.10)^{*}$ & $0.39(0.11)^{*}$ 
    & $0.62(0.12)^{*}$ & $0.40(0.09)^{*}$ \\
\bottomrule
\end{tabular}}
\caption{Group Avg. Similarity \textit{within} Coder Groups (standard deviation in brackets). Themes are considered to be more \textbf{cohesive} if intra-theme similarity is high and more \textbf{distinctive} if inter-theme similarity is low. \textbf{Bold} figures highlight more cohesive \textit{or} distinctive value between collaboration settings. $^*$Statistically significant using a two-sample unpaired t-test with $p<0.05$.}
\label{tab:cohesiveness}
\end{table*}


\begin{figure}[t]
    \centering
    \includegraphics[width=1\linewidth]{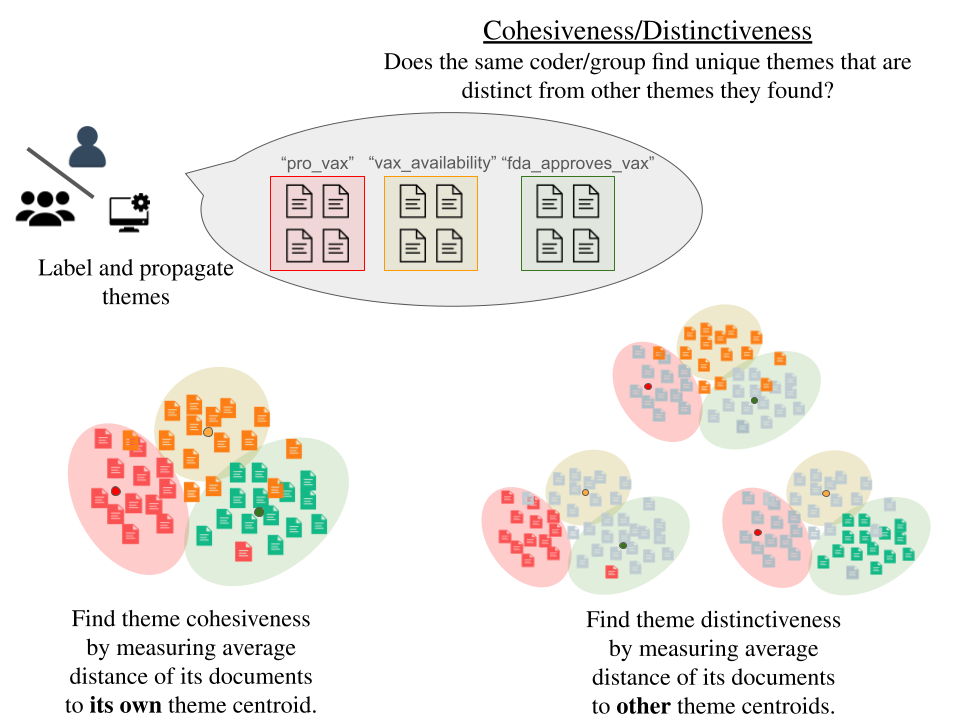}
    \captionsetup{singlelinecheck=off}
    \caption{Once a coder has identified themes and they have been propagated the full dataset, we calculate \textit{intra-theme similarity} by measuring the avg. of the pairwise distances between each document within a theme (left).
    We calculate \textit{inter-theme similarity} by measuring the avg. of pairwise distances between each document in a theme and documents assigned to all other themes (right)}
    \label{fig:cohesiveness}
\end{figure}

\subsection{Cohesiveness and Distinctiveness}
Another dimension for determining the systematicity and clarity of coding outcomes is by evaluating the similarities and differences between themes within the same code-book. We propose two metrics to measure this: cohesiveness and distinctiveness. A theme is said to be cohesive if its documents are similar to each other (measured by \textit{intra}-theme similarity) and distinctive if it is dissimilar from documents in other themes within the same code-book (measured by \textit{inter}-theme similarity). Intuitively, the purpose of grouping documents by theme is to create abstract representations of a dataset, where each theme represents a distinct facet of the data. If themes are not cohesive and distinctive, then it becomes hard to tell which theme a given document should belong to and the code-book falls apart.

Figure \ref{fig:cohesiveness} shows how to evaluate these metrics for a single coder (or coder group). We calculate both the intra-theme similarity and the inter-theme similarity for all the themes in the code-book. \textit{Intra-theme similarity} is calculated by taking the average of pair-wise similarity between all documents of the same theme. \textit{Inter-theme similarity} for a given theme is calculated by taking the average pair-wise similarity of documents in that theme with documents in all other themes. 

A confounding factor in these measures is that all systems provide broad coverage of documents such that even distantly related documents may be assigned to a theme. To more accurately represent the cohesiveness and distinctiveness of the themes in each experiment, we perform the same calculations on a subset comprised of only the top 25\% of documents most closely related to each theme. For the relational and LLM-based systems, this top quartile is selected using the distance from the centroid. For the interactive topic model, we use the weights assigned by the model.



Table \ref{tab:cohesiveness} shows results for both the whole dataset as well as the subset of the documents closest to each theme. Overall, we find that the intra-theme similarities are always higher than inter-theme similarities, which means that themes are at least moderately cohesive and distinctive across the board. For the COVID dataset, we find that themes may be more cohesive but not more distinctive in the synchronous setting especially for the relational system. This may be due to the homogeneity of the dataset, which further explains the uniformity of results in the topic model and LLM experiments. 

The results from the Climate dataset present a more compelling finding, especially since all differences are statistically significant, with the synchronous experiments uniformly producing much more cohesive themes. When only the top 25\% of documents are considered, the increase in cohesiveness is not correlated with a decrease in distinctiveness, unlike in the COVID case. This is best exemplified by the topic model results, where the synchronous experiment achieving 0.72 on intra-theme similarity, much higher than the 0.65 achieved by the asynchronous experiment despite both experiments producing the same inter-theme similarity. Our results strongly suggest that synchronous collaboration facilitates deeper analysis to uncover clearer themes in the data. This is further supported by the greater distinctiveness in the Climate data experiments: we expect coders to find more distinct themes in a more complex and heterogenous dataset.



\subsection{Correctness} Interactive systems allow users to automate large portions of the coding process at the risk of producing inaccurate theme assignments. To estimate how correct the outputs of each system are, we conduct a post-hoc analysis by manually checking a randomly selected sample of $2,400$ document-theme pairs ($200$ per experimental setting). To ensure the representativeness of our samples, we split the data into quartiles based on document relatedness to each theme and select a uniform sample of documents per theme. As before, relatedness is calculated using the theme weight distribution for the interactive topic model and distance from the centroid for the the other two systems. To assess reliability, each assignment is evaluated by two annotators, with a third weighing in for tie breaks. The human evaluators demonstrate moderate-to-high agreement, with a Krippendorff's $\alpha$ of $0.632$ for the COVID tweets dataset and $0.696$ for the Climate ads dataset, suggesting that we can trust these estimations.

\begin{figure}[t]
    \centering
    \includegraphics[width=1\linewidth]{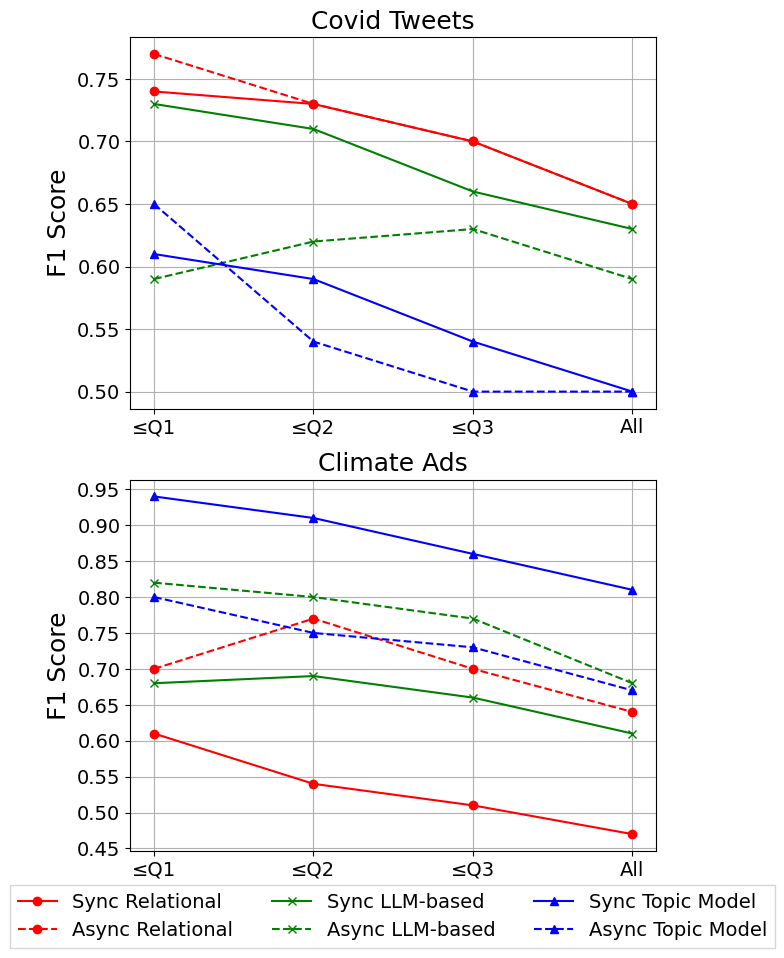}
    \caption{Correctness w.r.t. distance from theme.}
    \label{fig:correctness}
\end{figure}

Figure \ref{fig:correctness} shows the correctness results for each quartile sample per experiment. For the COVID dataset, we observe that the relational system is the most accurate and shows negligible differences across collaboration settings. This an encouraging result, given that this system took the most advantage of synchronous deliberation based on our other metrics of quality. The other two systems showed marked differences, with the topic modeling approach producing more accurate assignments in asynchronous operation, and the LLM producing more accurate assignment in a synchronous paradigm.

On the other hand, the Climate dataset shows almost the opposite result, with the synchronous relational experiment producing the least accurate results and the synchronous topic model significantly outperforming all the other experiments. While there is no strong relationship between collaboration setting and correctness scores, our findings suggest that dataset characteristics may significantly impact the performance of interactive systems. The Climate dataset has almost double the number of words per document compared to the COVID dataset (53.1 vs. 27.5), which may allow the topic model to capture more meaningful lexical statistics. The Climate dataset is also only a fraction of the size of the COVID dataset (5,471 documents vs. 85,799 documents), which means that topic model users see a much greater proportion of the documents and make more accurate refinement decisions.



\subsection{Impact of Dataset Differences}\label{sec:dataset_differences}

To investigate how structural and semantic characteristics of a dataset affect the coding process, we carried out a comprehensive analysis across multiple dimensions, including readability, templatic or formulaic patterns, semantic theme ambiguity, and semantic similarity distribution (details in Appendix~\ref{app:dataset}). The higher formulaic content in climate ads (20.1\% vs.\ 3.7\%) may facilitate easy pattern recognition, but could also induce coder fatigue through repetitive content. The greater linguistic complexity (Flesch Reading Ease 47.4 vs.\ 54.9; documents averaging 53.1 vs.\ 27.5 words) and semantic diversity (mean pairwise similarity 0.26 vs.\ 0.43) of climate ads likely impose higher cognitive load per coding decision.

Conversely, the semantic homogeneity of COVID tweets, evidenced by higher pairwise similarity (0.43 vs.\ 0.26) and centroid proximity (0.65 vs.\ 0.51), gives coders shared semantic reference points that can support more efficient real-time discussion in synchronous settings. The less formulaic nature of COVID discourse may further sustain coder engagement through content variety. These dynamics are consistent with our finding that average consistency scores favor synchronous groups for COVID (Table~\ref{tab:consistency}), while for climate ads, asynchronous coders perform on par with or slightly above synchronous groups.
Together, these results suggest that coding protocol selection should account for corpus-specific characteristics: semantically diverse, linguistically complex corpora may benefit from asynchronous approaches that allow extended reflection, while homogeneous corpora can be efficiently processed through synchronous collaborative coding.
\section{User Study and Recommendations}\label{sec:user_study}

We conducted semi-structured interviews to understand the participant experiences with the task and tools, with a focus on synchronous versus asynchronous coding (see Appendix~\ref{app:interview} for interview script). Several findings emerged.

\paragraph{Synchronous teamwork eased coding and improved outcomes.} 
Participants reported that working synchronously helped them contextualize data, resolve disagreements quickly, and ``break ties'' through discussion. These experiences align with our quantitative findings showing higher consistency and cohesiveness in synchronous settings, suggesting the value of systems that explicitly support real-time deliberation.

\paragraph{Asynchronous coders were more sensitive to tool limitations.} Because they worked largely in isolation, asynchronous participants focused more on usability issues and tool constraints, highlighting the need to improve support for independent coding workflows.

\paragraph{Limited control reduced user trust, particularly in topic modeling.} Participants reported a loss of agency when using the topic modeling system, citing insufficient control over operations and difficulty tracking theme evolution -- \textit{`the merge process did not offer the ideal amount of control and made it difficult to keep track of the theme groups.''}

Initial topics also sometimes conflated opposing themes due to lexical similarity, frustrating users' attempts to refine results -- \textit{``Many Anti-Vax and Pro-Vax standpoints use the same words/phrases in their tweets, which the \citet{fang-2023-user} model groups together despite the stark difference in message between the two.''} 

While some appreciated the model's initial theme induction, participants desired greater control and clearer explanations of cluster structure. These findings highlight the need for systems to maximize the degree of control afforded to users.


\paragraph{LLM-based coding was costly and unreliable at scale.} Despite strong reasoning capabilities, LLMs proved inefficient for large-scale annotation due to high computational costs and inconsistent classification performance. This opens an opportunity for NLP researchers to make LLMs more reliable inductive reasoners and to come up with prompting strategies that can allow LLMs to reliably classify documents in bulk, especially when working at scale. 

\paragraph{NLP researchers should consider ideal collaboration settings and dataset characteristics when designing tools.} Our results show that these factors have outsized impact on coding outcomes, and that no single configuration works best across conditions. Interactive systems should therefore balance automation with user control and offer tailored workflows that support both real-time deliberation and independent coding at scale.


\section{Conclusion and Future Work}

We examined three categories of NLP-assisted qualitative research tools across synchronous and asynchronous collaboration settings, applying them to two structurally distinct English corpora. We designed an evaluation framework that captures theme consistency, cohesiveness, distinctiveness, and correctness, and used it to analyze how coding outcomes vary across methods, settings, and data.

We find that collaboration modality significantly affects output quality, but the direction of the effect depends on both the system and the dataset. On homogeneous corpora, synchronous collaboration produces more consistent and cohesive themes. On heterogeneous corpora, synchronous coding yields markedly more cohesive themes without sacrificing distinctiveness, while consistency scores are comparable across settings. We further observe that systems affording richer user intervention benefit more from synchronous deliberation, whereas systems with limited intervention show little sensitivity to collaboration setting. 

Beyond collaboration, we find that dataset characteristics independently shape system performance, as the same system can be the strongest or among the weakest depending on the corpus. Topic modeling, for instance, achieves the highest correctness on Climate ads in the synchronous setting but produces highly divergent code-books across asynchronous coders on COVID. LLM-based solutions show promise but remain costly and inconsistent at scale, and offer fewer intervention points for coders to leverage deliberation.

While this study focuses on collaboration modalities, there are numerous other variables that can affect a tool's efficacy for qualitative coding. We believe that our proposed evaluation framework can be repurposed and expanded to evaluate a wide range of interventions, such as the underlying NLP technology, the interactive interface, the expertise of the coders, and the type of data being annotated. Future work should focus on evaluating other interactive systems to determine whether our findings are representative of relationships between these variables. We hope to inform the development of more robust evaluations of NLP tools for qualitative research in realistic settings.
\section*{Limitations}

The study presented in this paper has two main limitations.

(1) While we selected three distinct, representative tools to perform our analysis of synchronous vs. asynchronous settings, as well as two datasets with distict characteristics, the list is of course non-exhaustive. A larger study incorporating more tools and datasets could yield additional insights.



(2) While we look at an important variable in qualitative research settings (collaboration modality), there are several other variables that can influence the outcome of NLP-assisted solutions (e.g., choice of tool, expertise and live experience of annotators, type of data being annotated, etc.). In addition to this, we did not explore the many different consolidation strategies that are often used to bring together the perspectives of asynchronous coders. We leave the explorations of these questions for future work. 
\section*{Ethical Considerations}

To the best of our knowledge, no code of ethics was violated during the development of this project. We used publicly available tools and datasets according to their licensing agreements. For our annotation experiments, we followed IRB protocol and did not retain any personally identifiable information. 

All information needed to replicate our experiments is presented in the paper. We reported all experimental settings, as well as any pre-processing steps, learning configurations, hyper-parameters, and additional technical details. Due to space constraints, some of this information was relegated to the Appendix. In addition to this, we have made the results of the annotation experiment available to the community, as well as the code to produce all of our reported results. We believe that the results reported in this paper support our claims. 

\section*{Acknowledgments}

This work utilized computing resources at the University of Colorado Boulder, including the CUmulus on-premise cloud service and the Blanca condo computing resource. CUmulus is jointly funded by the National Science Foundation (award OAC-1925766) and the University of Colorado Boulder. Blanca is jointly funded by its computing users and the University of Colorado Boulder.

\bibliography{anthology,custom}

@article{Baden02012022,
author = {Christian Baden and Christian Pipal and Martijn Schoonvelde and Mariken A. C. G van der Velden},
title = {Three Gaps in Computational Text Analysis Methods for Social Sciences: A Research Agenda},
journal = {Communication Methods and Measures},
volume = {16},
number = {1},
pages = {1--18},
year = {2022},
publisher = {Routledge},
doi = {10.1080/19312458.2021.2015574},
URL = { 
        https://doi.org/10.1080/19312458.2021.2015574
},
eprint = { 
        https://doi.org/10.1080/19312458.2021.2015574
}
}

@misc{gao-2024-collabcoder,
    title={CollabCoder: A Lower-barrier, Rigorous Workflow for Inductive Collaborative Qualitative Analysis with Large Language Models}, 
    author={Jie Gao and Yuchen Guo and Gionnieve Lim and Tianqin Zhang and Zheng Zhang and Toby Jia-Jun Li and Simon Tangi Perrault},
    year={2024},
    eprint={2304.07366},
    archivePrefix={arXiv},
    primaryClass={cs.HC}
}

@inproceedings{fang-2023-user,
    title = "A User-Centered, Interactive, Human-in-the-Loop Topic Modelling System",
    author = "Fang, Zheng  and
    Alqazlan, Lama  and
    Liu, Du  and
    He, Yulan  and
    Procter, Rob",
    editor = "Vlachos, Andreas  and
    Augenstein, Isabelle",
    booktitle = "Proceedings of the 17th Conference of the European Chapter of the Association for Computational Linguistics",
    month = may,
    year = "2023",
    address = "Dubrovnik, Croatia",
    publisher = "Association for Computational Linguistics",
    url = "https://aclanthology.org/2023.eacl-main.37",
    doi = "10.18653/v1/2023.eacl-main.37",
    pages = "505--522",
}

@inproceedings{pacheco-2023-interactive,
    title = "Interactive Concept Learning for Uncovering Latent Themes in Large Text Collections",
    author = "Pacheco, Maria Leonor  and
      Islam, Tunazzina  and
      Ungar, Lyle  and
      Yin, Ming  and
      Goldwasser, Dan",
    editor = "Rogers, Anna  and
      Boyd-Graber, Jordan  and
      Okazaki, Naoaki",
    booktitle = "Findings of the Association for Computational Linguistics: ACL 2023",
    month = jul,
    year = "2023",
    address = "Toronto, Canada",
    publisher = "Association for Computational Linguistics",
    url = "https://aclanthology.org/2023.findings-acl.313",
    doi = "10.18653/v1/2023.findings-acl.313",
    pages = "5059--5080",
}

@inproceedings{smith-2018-loop,
    author = {Smith, Alison and Kumar, Varun and Boyd-Graber, Jordan and Seppi, Kevin and Findlater, Leah},
    title = {Closing the Loop: User-Centered Design and Evaluation of a Human-in-the-Loop Topic Modeling System},
    year = {2018},
    isbn = {9781450349451},
    publisher = {Association for Computing Machinery},
    address = {New York, NY, USA},
    url = {https://doi.org/10.1145/3172944.3172965},
    doi = {10.1145/3172944.3172965},
    booktitle = {Proceedings of the 23rd International Conference on Intelligent User Interfaces},
    pages = {293–304},
    numpages = {12},
    location = {<conf-loc>, <city>Tokyo</city>, <country>Japan</country>, </conf-loc>},
    series = {IUI '18}
}

@article{gao-2023-coaicoder,
    author = {Gao, Jie and Choo, Kenny Tsu Wei and Cao, Junming and Lee, Roy Ka-Wei and Perrault, Simon},
    title = {CoAIcoder: Examining the Effectiveness of AI-assisted Human-to-Human Collaboration in Qualitative Analysis},
    year = {2023},
    issue_date = {February 2024},
    publisher = {Association for Computing Machinery},
    address = {New York, NY, USA},
    volume = {31},
    number = {1},
    issn = {1073-0516},
    url = {https://doi.org/10.1145/3617362},
    doi = {10.1145/3617362},
    journal = {ACM Trans. Comput.-Hum. Interact.},
    month = {nov},
    articleno = {6},
    numpages = {38}
}

@inproceedings{zade-2018-disagree,
    author = {Zade, Himanshu and Drouhard, Margaret and Chinh, Bonnie and Gan, Lu and Aragon, Cecilia},
    title = {Conceptualizing Disagreement in Qualitative Coding},
    year = {2018},
    isbn = {9781450356206},
    publisher = {Association for Computing Machinery},
    address = {New York, NY, USA},
    url = {https://doi.org/10.1145/3173574.3173733},
    doi = {10.1145/3173574.3173733},
    booktitle = {Proceedings of the 2018 CHI Conference on Human Factors in Computing Systems},
    pages = {1–11},
    numpages = {11},
    location = {<conf-loc>, <city>Montreal QC</city>, <country>Canada</country>, </conf-loc>},
    series = {CHI '18}
}

@inbook{cornish-2014-sage-cqa,
    address = {London; London},
    author = {Cornish, Flora and Gillespie, Alex and Zittoun, Tania},
    c1 = {The SAGE Handbook of Qualitative Data Analysis},
    c2 = {pages 79-93},
    date = {2024/04/07},
    doi = {10.4135/9781446282243},
    publisher = {SAGE Publications Ltd},
    title = {The SAGE Handbook of Qualitative Data Analysis},
    url = {https://methods.sagepub.com/book/the-sage-handbook-of-qualitative-data-analysis},
    year = {2014}
}

@article{braun-2006,
	author = {Braun, Virginia and Clarke, Victoria},
	doi = {10.1191/1478088706qp063oa},
	journal = {Qualitative Research in Psychology},
	month = {01},
	pages = {77-101},
	title = {Using thematic analysis in psychology},
	volume = {3},
	year = {2006}
}

@article{glaser1968discovery,
    title={The discovery of grounded theory; strategies for qualitative research},
    author={Glaser, Barney G and Strauss, Anselm L and Strutzel, Elizabeth},
    journal={Nursing research},
    volume={17},
    number={4},
    pages={364},
    year={1968},
    publisher={LWW}
}

@article{jiang-2021-serendipity,
    author = {Jiang, Jialun Aaron and Wade, Kandrea and Fiesler, Casey and Brubaker, Jed R.},
    title = {Supporting Serendipity: Opportunities and Challenges for Human-AI Collaboration in Qualitative Analysis},
    year = {2021},
    issue_date = {April 2021},
    publisher = {Association for Computing Machinery},
    address = {New York, NY, USA},
    volume = {5},
    number = {CSCW1},
    url = {https://doi.org/10.1145/3449168},
    doi = {10.1145/3449168},
    journal = {Proc. ACM Hum.-Comput. Interact.},
    month = {apr},
    articleno = {94},
    numpages = {23}
}

@article{feuston2021tools,
    author = {Feuston, Jessica L. and Brubaker, Jed R.},
    title = {Putting Tools in Their Place: The Role of Time and Perspective in Human-AI Collaboration for Qualitative Analysis},
    year = {2021},
    issue_date = {October 2021},
    publisher = {Association for Computing Machinery},
    address = {New York, NY, USA},
    volume = {5},
    number = {CSCW2},
    url = {https://doi.org/10.1145/3479856},
    doi = {10.1145/3479856},
    journal = {Proc. ACM Hum.-Comput. Interact.},
    month = {oct},
    articleno = {469},
    numpages = {25}
}

@article{chen-2018-ambiguity,
    author = {Chen, Nan-Chen and Drouhard, Margaret and Kocielnik, Rafal and Suh, Jina and Aragon, Cecilia R.},
    title = {Using Machine Learning to Support Qualitative Coding in Social Science: Shifting the Focus to Ambiguity},
    year = {2018},
    issue_date = {June 2018},
    publisher = {Association for Computing Machinery},
    address = {New York, NY, USA},
    volume = {8},
    number = {2},
    issn = {2160-6455},
    url = {https://doi.org/10.1145/3185515},
    doi = {10.1145/3185515},
    journal = {ACM Trans. Interact. Intell. Syst.},
    month = {jun},
    articleno = {9},
    numpages = {20}
}

@inproceedings{dai-etal-2023-llm,
    title = "{LLM}-in-the-loop: Leveraging Large Language Model for Thematic Analysis",
    author = "Dai, Shih-Chieh  and
      Xiong, Aiping  and
      Ku, Lun-Wei",
    editor = "Bouamor, Houda  and
      Pino, Juan  and
      Bali, Kalika",
    booktitle = "Findings of the Association for Computational Linguistics: EMNLP 2023",
    month = dec,
    year = "2023",
    address = "Singapore",
    publisher = "Association for Computational Linguistics",
    url = "https://aclanthology.org/2023.findings-emnlp.669",
    doi = "10.18653/v1/2023.findings-emnlp.669",
    pages = "9993--10001",
}

@article{doi:10.1146/annurev-polisci-090216-023229,
author = {Brady, Henry E.},
title = {The Challenge of Big Data and Data Science},
journal = {Annual Review of Political Science},
volume = {22},
number = {1},
pages = {297-323},
year = {2019},
doi = {10.1146/annurev-polisci-090216-023229},
URL = {https://doi.org/10.1146/annurev-polisci-090216-023229},
eprint = {https://doi.org/10.1146/annurev-polisci-090216-023229}}

@article{IJoC10675,
	author = {Martin Hilbert and George Barnett and Joshua Blumenstock and Noshir Contractor and Jana Diesner and Seth Frey and Sandra González-Bailón and PJ Lamberson and Jennifer Pan and Tai-Quan Peng and Cuihua (Cindy) Shen and Paul E. Smaldino and Wouter van Atteveldt and Annie Waldherr and Jingwen Zhang and Jonathan J. H. Zhu},
	title = {Computational Communication Science| Computational Communication Science: A Methodological Catalyzer for a Maturing Discipline},
	journal = {International Journal of Communication},
	volume = {13},
	number = {0},
	year = {2019},
	issn = {1932-8036},	url = {https://ijoc.org/index.php/ijoc/article/view/10675}
}

@article{cite-key,
	Address = {London},
	Author = {Flick, Uwe},
	Da = {2024/08/14},
	Doi = {10.4135/9781446282243},
	Title = {The SAGE Handbook of Qualitative Data Analysis},
	Ty = {EBOOK},
	Url = {https://methods.sagepub.com/book/the-sage-handbook-of-qualitative-data-analysis},
	Year = {2014}}

@inproceedings{pacheco-etal-2022-holistic,
    title = "A Holistic Framework for Analyzing the {COVID}-19 Vaccine Debate",
    author = "Pacheco, Maria Leonor  and
      Islam, Tunazzina  and
      Mahajan, Monal  and
      Shor, Andrey  and
      Yin, Ming  and
      Ungar, Lyle  and
      Goldwasser, Dan",
    editor = "Carpuat, Marine  and
      de Marneffe, Marie-Catherine  and
      Meza Ruiz, Ivan Vladimir",
    booktitle = "Proceedings of the 2022 Conference of the North American Chapter of the Association for Computational Linguistics: Human Language Technologies",
    month = jul,
    year = "2022",
    address = "Seattle, United States",
    publisher = "Association for Computational Linguistics",
    url = "https://aclanthology.org/2022.naacl-main.427",
    doi = "10.18653/v1/2022.naacl-main.427",
    pages = "5821--5839",
}

@inproceedings{roy-etal-2021-identifying,
    title = "Identifying Morality Frames in Political Tweets using Relational Learning",
    author = "Roy, Shamik  and
      Pacheco, Maria Leonor  and
      Goldwasser, Dan",
    editor = "Moens, Marie-Francine  and
      Huang, Xuanjing  and
      Specia, Lucia  and
      Yih, Scott Wen-tau",
    booktitle = "Proceedings of the 2021 Conference on Empirical Methods in Natural Language Processing",
    month = nov,
    year = "2021",
    address = "Online and Punta Cana, Dominican Republic",
    publisher = "Association for Computational Linguistics",
    url = "https://aclanthology.org/2021.emnlp-main.783",
    doi = "10.18653/v1/2021.emnlp-main.783",
    pages = "9939--9958",
}

@INPROCEEDINGS{drouhard-aeonium,
    author={Drouhard, Margaret and Chen, Nan-Chen and Suh, Jina and Kocielnik, Rafal and Peña-Araya, Vanessa and Cen, Keting and Xiangyi Zheng and Aragon, Cecilia R.},
    booktitle={2017 IEEE Pacific Visualization Symposium (PacificVis)}, 
    title={Aeonium: Visual analytics to support collaborative qualitative coding}, 
    year={2017},
    volume={},
    number={},
    pages={220-229},
    doi={10.1109/PACIFICVIS.2017.8031598}
}

@inproceedings{fang-etal-2021-query,
    title = "A Query-Driven Topic Model",
    author = "Fang, Zheng  and
      He, Yulan  and
      Procter, Rob",
    editor = "Zong, Chengqing  and
      Xia, Fei  and
      Li, Wenjie  and
      Navigli, Roberto",
    booktitle = "Findings of the Association for Computational Linguistics: ACL-IJCNLP 2021",
    month = aug,
    year = "2021",
    address = "Online",
    publisher = "Association for Computational Linguistics",
    url = "https://aclanthology.org/2021.findings-acl.154",
    doi = "10.18653/v1/2021.findings-acl.154",
    pages = "1764--1777",
}

@ARTICLE{choo-utopian,

  author={Choo, Jaegul and Lee, Changhyun and Reddy, Chandan K. and Park, Haesun},

  journal={IEEE Transactions on Visualization and Computer Graphics}, 

  title={UTOPIAN: User-Driven Topic Modeling Based on Interactive Nonnegative Matrix Factorization}, 

  year={2013},

  volume={19},

  number={12},

  pages={1992-2001},

  doi={10.1109/TVCG.2013.212}
}

@article{hoque-convisit,
    author = {Hoque, Enamul and Carenini, Giuseppe},
    title = {Interactive Topic Modeling for Exploring Asynchronous Online Conversations: Design and Evaluation of ConVisIT},
    year = {2016},
    issue_date = {May 2016},
    publisher = {Association for Computing Machinery},
    address = {New York, NY, USA},
    volume = {6},
    number = {1},
    issn = {2160-6455},
    url = {https://doi.org/10.1145/2854158},
    doi = {10.1145/2854158},
    journal = {ACM Trans. Interact. Intell. Syst.},
    month = {feb},
    articleno = {7},
    numpages = {24}
}

@inproceedings{rietz-cody,
    author = {Rietz, Tim and Toreini, Peyman and Maedche, Alexander},
    title = {Cody: An Interactive Machine Learning System for Qualitative Coding},
    year = {2020},
    isbn = {9781450375153},
    publisher = {Association for Computing Machinery},
    address = {New York, NY, USA},
    url = {https://doi.org/10.1145/3379350.3416195},
    doi = {10.1145/3379350.3416195},
    booktitle = {Adjunct Proceedings of the 33rd Annual ACM Symposium on User Interface Software and Technology},
    pages = {90–92},
    numpages = {3},
    location = {Virtual Event, USA},
    series = {UIST '20 Adjunct}
}

@inproceedings{NIPS2008_beed1360,
 author = {Ben-David, Shai and Ackerman, Margareta},
 booktitle = {Advances in Neural Information Processing Systems},
 editor = {D. Koller and D. Schuurmans and Y. Bengio and L. Bottou},
 pages = {},
 publisher = {Curran Associates, Inc.},
 title = {Measures of Clustering Quality: A Working Set of Axioms for Clustering},
 url = {https://proceedings.neurips.cc/paper_files/paper/2008/file/beed13602b9b0e6ecb5b568ff5058f07-Paper.pdf},
 volume = {21},
 year = {2008}
}

@inproceedings{reimers-gurevych-2019-sentence,
    title = "Sentence-{BERT}: Sentence Embeddings using {S}iamese {BERT}-Networks",
    author = "Reimers, Nils  and
      Gurevych, Iryna",
    editor = "Inui, Kentaro  and
      Jiang, Jing  and
      Ng, Vincent  and
      Wan, Xiaojun",
    booktitle = "Proceedings of the 2019 Conference on Empirical Methods in Natural Language Processing and the 9th International Joint Conference on Natural Language Processing (EMNLP-IJCNLP)",
    month = nov,
    year = "2019",
    address = "Hong Kong, China",
    publisher = "Association for Computational Linguistics",
    url = "https://aclanthology.org/D19-1410",
    doi = "10.18653/v1/D19-1410",
    pages = "3982--3992",
}

@inproceedings{10.5555/3540261.3540416,
author = {Hoyle, Alexander and Goel, Pranav and Peskov, Denis and Hian-Cheong, Andrew and Boyd-Graber, Jordan and Resnik, Philip},
title = {Is automated topic model evaluation broken? the incoherence of coherence},
year = {2021},
isbn = {9781713845393},
publisher = {Curran Associates Inc.},
address = {Red Hook, NY, USA},
booktitle = {Proceedings of the 35th International Conference on Neural Information Processing Systems},
articleno = {155},
numpages = {16},
series = {NIPS '21}
}

@inproceedings{llms0shotreasoners,
author = {Kojima, Takeshi and Gu, Shixiang Shane and Reid, Machel and Matsuo, Yutaka and Iwasawa, Yusuke},
title = {Large language models are zero-shot reasoners},
year = {2024},
isbn = {9781713871088},
publisher = {Curran Associates Inc.},
address = {Red Hook, NY, USA},
booktitle = {Proceedings of the 36th International Conference on Neural Information Processing Systems},
articleno = {1613},
numpages = {15},
location = {New Orleans, LA, USA},
series = {NIPS '22}
}

@inproceedings{deductivecodingwithLLM,
author = {Xiao, Ziang and Yuan, Xingdi and Liao, Q. Vera and Abdelghani, Rania and Oudeyer, Pierre-Yves},
title = {Supporting Qualitative Analysis with Large Language Models: Combining Codebook with GPT-3 for Deductive Coding},
year = {2023},
isbn = {9798400701078},
publisher = {Association for Computing Machinery},
address = {New York, NY, USA},
url = {https://doi.org/10.1145/3581754.3584136},
doi = {10.1145/3581754.3584136},
booktitle = {Companion Proceedings of the 28th International Conference on Intelligent User Interfaces},
pages = {75–78},
numpages = {4},
location = {Sydney, NSW, Australia},
series = {IUI '23 Companion}
}

@misc{chew2023llmassistedcontentanalysisusing,
      title={LLM-Assisted Content Analysis: Using Large Language Models to Support Deductive Coding}, 
      author={Robert Chew and John Bollenbacher and Michael Wenger and Jessica Speer and Annice Kim},
      year={2023},
      eprint={2306.14924},
      archivePrefix={arXiv},
      primaryClass={cs.CL},
      url={https://arxiv.org/abs/2306.14924}, 
}

@article{gwet2008computing,
  title={Computing inter-rater reliability and its variance in the presence of high agreement},
  author={Gwet, Kilem Li},
  journal={British Journal of Mathematical and Statistical Psychology},
  volume={61},
  number={1},
  pages={29--48},
  year={2008},
  publisher={Wiley Online Library}
}

@inproceedings{Chuang2013DocumentEW,
  title={Document Exploration with Topic Modeling : Designing Interactive Visualizations to Support Effective Analysis Workflows},
  author={Jason Chuang and Daniel A. McFarland},
  year={2013},
  url={https://api.semanticscholar.org/CorpusID:43940920}
}

@misc{grattafiori2024llama3herdmodels,
      title={The Llama 3 Herd of Models}, 
      author={Aaron Grattafiori and Abhimanyu Dubey and Abhinav Jauhri and Abhinav Pandey and Abhishek Kadian and Ahmad Al-Dahle and Aiesha Letman and Akhil Mathur and Alan Schelten and Alex Vaughan and Amy Yang and Angela Fan and Anirudh Goyal and Anthony Hartshorn and Aobo Yang and Archi Mitra and Archie Sravankumar and Artem Korenev and Arthur Hinsvark and Arun Rao and Aston Zhang and Aurelien Rodriguez and Austen Gregerson and Ava Spataru and Baptiste Roziere and Bethany Biron and Binh Tang and Bobbie Chern and Charlotte Caucheteux and Chaya Nayak and Chloe Bi and Chris Marra and Chris McConnell and Christian Keller and Christophe Touret and Chunyang Wu and Corinne Wong and Cristian Canton Ferrer and Cyrus Nikolaidis and Damien Allonsius and Daniel Song and Danielle Pintz and Danny Livshits and Danny Wyatt and David Esiobu and Dhruv Choudhary and Dhruv Mahajan and Diego Garcia-Olano and Diego Perino and Dieuwke Hupkes and Egor Lakomkin and Ehab AlBadawy and Elina Lobanova and Emily Dinan and Eric Michael Smith and Filip Radenovic and Francisco Guzmán and Frank Zhang and Gabriel Synnaeve and Gabrielle Lee and Georgia Lewis Anderson and Govind Thattai and Graeme Nail and Gregoire Mialon and Guan Pang and Guillem Cucurell and Hailey Nguyen and Hannah Korevaar and Hu Xu and Hugo Touvron and Iliyan Zarov and Imanol Arrieta Ibarra and Isabel Kloumann and Ishan Misra and Ivan Evtimov and Jack Zhang and Jade Copet and Jaewon Lee and Jan Geffert and Jana Vranes and Jason Park and Jay Mahadeokar and Jeet Shah and Jelmer van der Linde and Jennifer Billock and Jenny Hong and Jenya Lee and Jeremy Fu and Jianfeng Chi and Jianyu Huang and Jiawen Liu and Jie Wang and Jiecao Yu and Joanna Bitton and Joe Spisak and Jongsoo Park and Joseph Rocca and Joshua Johnstun and Joshua Saxe and Junteng Jia and Kalyan Vasuden Alwala and Karthik Prasad and Kartikeya Upasani and Kate Plawiak and Ke Li and Kenneth Heafield and Kevin Stone and Khalid El-Arini and Krithika Iyer and Kshitiz Malik and Kuenley Chiu and Kunal Bhalla and Kushal Lakhotia and Lauren Rantala-Yeary and Laurens van der Maaten and Lawrence Chen and Liang Tan and Liz Jenkins and Louis Martin and Lovish Madaan and Lubo Malo and Lukas Blecher and Lukas Landzaat and Luke de Oliveira and Madeline Muzzi and Mahesh Pasupuleti and Mannat Singh and Manohar Paluri and Marcin Kardas and Maria Tsimpoukelli and Mathew Oldham and Mathieu Rita and Maya Pavlova and Melanie Kambadur and Mike Lewis and Min Si and Mitesh Kumar Singh and Mona Hassan and Naman Goyal and Narjes Torabi and Nikolay Bashlykov and Nikolay Bogoychev and Niladri Chatterji and Ning Zhang and Olivier Duchenne and Onur Çelebi and Patrick Alrassy and Pengchuan Zhang and Pengwei Li and Petar Vasic and Peter Weng and Prajjwal Bhargava and Pratik Dubal and Praveen Krishnan and Punit Singh Koura and Puxin Xu and Qing He and Qingxiao Dong and Ragavan Srinivasan and Raj Ganapathy and Ramon Calderer and Ricardo Silveira Cabral and Robert Stojnic and Roberta Raileanu and Rohan Maheswari and Rohit Girdhar and Rohit Patel and Romain Sauvestre and Ronnie Polidoro and Roshan Sumbaly and Ross Taylor and Ruan Silva and Rui Hou and Rui Wang and Saghar Hosseini and Sahana Chennabasappa and Sanjay Singh and Sean Bell and Seohyun Sonia Kim and Sergey Edunov and Shaoliang Nie and Sharan Narang and Sharath Raparthy and Sheng Shen and Shengye Wan and Shruti Bhosale and Shun Zhang and Simon Vandenhende and Soumya Batra and Spencer Whitman and Sten Sootla and Stephane Collot and Suchin Gururangan and Sydney Borodinsky and Tamar Herman and Tara Fowler and Tarek Sheasha and Thomas Georgiou and Thomas Scialom and Tobias Speckbacher and Todor Mihaylov and Tong Xiao and Ujjwal Karn and Vedanuj Goswami and Vibhor Gupta and Vignesh Ramanathan and Viktor Kerkez and Vincent Gonguet and Virginie Do and Vish Vogeti and Vítor Albiero and Vladan Petrovic and Weiwei Chu and Wenhan Xiong and Wenyin Fu and Whitney Meers and Xavier Martinet and Xiaodong Wang and Xiaofang Wang and Xiaoqing Ellen Tan and Xide Xia and Xinfeng Xie and Xuchao Jia and Xuewei Wang and Yaelle Goldschlag and Yashesh Gaur and Yasmine Babaei and Yi Wen and Yiwen Song and Yuchen Zhang and Yue Li and Yuning Mao and Zacharie Delpierre Coudert and Zheng Yan and Zhengxing Chen and Zoe Papakipos and Aaditya Singh and Aayushi Srivastava and Abha Jain and Adam Kelsey and Adam Shajnfeld and Adithya Gangidi and Adolfo Victoria and Ahuva Goldstand and Ajay Menon and Ajay Sharma and Alex Boesenberg and Alexei Baevski and Allie Feinstein and Amanda Kallet and Amit Sangani and Amos Teo and Anam Yunus and Andrei Lupu and Andres Alvarado and Andrew Caples and Andrew Gu and Andrew Ho and Andrew Poulton and Andrew Ryan and Ankit Ramchandani and Annie Dong and Annie Franco and Anuj Goyal and Aparajita Saraf and Arkabandhu Chowdhury and Ashley Gabriel and Ashwin Bharambe and Assaf Eisenman and Azadeh Yazdan and Beau James and Ben Maurer and Benjamin Leonhardi and Bernie Huang and Beth Loyd and Beto De Paola and Bhargavi Paranjape and Bing Liu and Bo Wu and Boyu Ni and Braden Hancock and Bram Wasti and Brandon Spence and Brani Stojkovic and Brian Gamido and Britt Montalvo and Carl Parker and Carly Burton and Catalina Mejia and Ce Liu and Changhan Wang and Changkyu Kim and Chao Zhou and Chester Hu and Ching-Hsiang Chu and Chris Cai and Chris Tindal and Christoph Feichtenhofer and Cynthia Gao and Damon Civin and Dana Beaty and Daniel Kreymer and Daniel Li and David Adkins and David Xu and Davide Testuggine and Delia David and Devi Parikh and Diana Liskovich and Didem Foss and Dingkang Wang and Duc Le and Dustin Holland and Edward Dowling and Eissa Jamil and Elaine Montgomery and Eleonora Presani and Emily Hahn and Emily Wood and Eric-Tuan Le and Erik Brinkman and Esteban Arcaute and Evan Dunbar and Evan Smothers and Fei Sun and Felix Kreuk and Feng Tian and Filippos Kokkinos and Firat Ozgenel and Francesco Caggioni and Frank Kanayet and Frank Seide and Gabriela Medina Florez and Gabriella Schwarz and Gada Badeer and Georgia Swee and Gil Halpern and Grant Herman and Grigory Sizov and Guangyi and Zhang and Guna Lakshminarayanan and Hakan Inan and Hamid Shojanazeri and Han Zou and Hannah Wang and Hanwen Zha and Haroun Habeeb and Harrison Rudolph and Helen Suk and Henry Aspegren and Hunter Goldman and Hongyuan Zhan and Ibrahim Damlaj and Igor Molybog and Igor Tufanov and Ilias Leontiadis and Irina-Elena Veliche and Itai Gat and Jake Weissman and James Geboski and James Kohli and Janice Lam and Japhet Asher and Jean-Baptiste Gaya and Jeff Marcus and Jeff Tang and Jennifer Chan and Jenny Zhen and Jeremy Reizenstein and Jeremy Teboul and Jessica Zhong and Jian Jin and Jingyi Yang and Joe Cummings and Jon Carvill and Jon Shepard and Jonathan McPhie and Jonathan Torres and Josh Ginsburg and Junjie Wang and Kai Wu and Kam Hou U and Karan Saxena and Kartikay Khandelwal and Katayoun Zand and Kathy Matosich and Kaushik Veeraraghavan and Kelly Michelena and Keqian Li and Kiran Jagadeesh and Kun Huang and Kunal Chawla and Kyle Huang and Lailin Chen and Lakshya Garg and Lavender A and Leandro Silva and Lee Bell and Lei Zhang and Liangpeng Guo and Licheng Yu and Liron Moshkovich and Luca Wehrstedt and Madian Khabsa and Manav Avalani and Manish Bhatt and Martynas Mankus and Matan Hasson and Matthew Lennie and Matthias Reso and Maxim Groshev and Maxim Naumov and Maya Lathi and Meghan Keneally and Miao Liu and Michael L. Seltzer and Michal Valko and Michelle Restrepo and Mihir Patel and Mik Vyatskov and Mikayel Samvelyan and Mike Clark and Mike Macey and Mike Wang and Miquel Jubert Hermoso and Mo Metanat and Mohammad Rastegari and Munish Bansal and Nandhini Santhanam and Natascha Parks and Natasha White and Navyata Bawa and Nayan Singhal and Nick Egebo and Nicolas Usunier and Nikhil Mehta and Nikolay Pavlovich Laptev and Ning Dong and Norman Cheng and Oleg Chernoguz and Olivia Hart and Omkar Salpekar and Ozlem Kalinli and Parkin Kent and Parth Parekh and Paul Saab and Pavan Balaji and Pedro Rittner and Philip Bontrager and Pierre Roux and Piotr Dollar and Polina Zvyagina and Prashant Ratanchandani and Pritish Yuvraj and Qian Liang and Rachad Alao and Rachel Rodriguez and Rafi Ayub and Raghotham Murthy and Raghu Nayani and Rahul Mitra and Rangaprabhu Parthasarathy and Raymond Li and Rebekkah Hogan and Robin Battey and Rocky Wang and Russ Howes and Ruty Rinott and Sachin Mehta and Sachin Siby and Sai Jayesh Bondu and Samyak Datta and Sara Chugh and Sara Hunt and Sargun Dhillon and Sasha Sidorov and Satadru Pan and Saurabh Mahajan and Saurabh Verma and Seiji Yamamoto and Sharadh Ramaswamy and Shaun Lindsay and Shaun Lindsay and Sheng Feng and Shenghao Lin and Shengxin Cindy Zha and Shishir Patil and Shiva Shankar and Shuqiang Zhang and Shuqiang Zhang and Sinong Wang and Sneha Agarwal and Soji Sajuyigbe and Soumith Chintala and Stephanie Max and Stephen Chen and Steve Kehoe and Steve Satterfield and Sudarshan Govindaprasad and Sumit Gupta and Summer Deng and Sungmin Cho and Sunny Virk and Suraj Subramanian and Sy Choudhury and Sydney Goldman and Tal Remez and Tamar Glaser and Tamara Best and Thilo Koehler and Thomas Robinson and Tianhe Li and Tianjun Zhang and Tim Matthews and Timothy Chou and Tzook Shaked and Varun Vontimitta and Victoria Ajayi and Victoria Montanez and Vijai Mohan and Vinay Satish Kumar and Vishal Mangla and Vlad Ionescu and Vlad Poenaru and Vlad Tiberiu Mihailescu and Vladimir Ivanov and Wei Li and Wenchen Wang and Wenwen Jiang and Wes Bouaziz and Will Constable and Xiaocheng Tang and Xiaojian Wu and Xiaolan Wang and Xilun Wu and Xinbo Gao and Yaniv Kleinman and Yanjun Chen and Ye Hu and Ye Jia and Ye Qi and Yenda Li and Yilin Zhang and Ying Zhang and Yossi Adi and Youngjin Nam and Yu and Wang and Yu Zhao and Yuchen Hao and Yundi Qian and Yunlu Li and Yuzi He and Zach Rait and Zachary DeVito and Zef Rosnbrick and Zhaoduo Wen and Zhenyu Yang and Zhiwei Zhao and Zhiyu Ma},
      year={2024},
      eprint={2407.21783},
      archivePrefix={arXiv},
      primaryClass={cs.AI},
      url={https://arxiv.org/abs/2407.21783}, 
}

@Inbook{Jin2010,
    author="Jin, Xin
    and Han, Jiawei",
    title="K-Means Clustering",
    bookTitle="Encyclopedia of Machine Learning",
    year="2010",
    publisher="Springer US",
    address="Boston, MA",
    pages="563--564",
    isbn="978-0-387-30164-8",
    doi="10.1007/978-0-387-30164-8_425",
    url="https://doi.org/10.1007/978-0-387-30164-8_425"
}

@article{JMLR:v9:vandermaaten08a,
  author  = {Laurens van der Maaten and Geoffrey Hinton},
  title   = {Visualizing Data using t-SNE},
  journal = {Journal of Machine Learning Research},
  year    = {2008},
  volume  = {9},
  number  = {86},
  pages   = {2579--2605},
  url     = {http://jmlr.org/papers/v9/vandermaaten08a.html}
}

@inproceedings{mendelsohn-etal-2021-modeling,
    title = "Modeling Framing in Immigration Discourse on Social Media",
    author = "Mendelsohn, Julia  and
      Budak, Ceren  and
      Jurgens, David",
    editor = "Toutanova, Kristina  and
      Rumshisky, Anna  and
      Zettlemoyer, Luke  and
      Hakkani-Tur, Dilek  and
      Beltagy, Iz  and
      Bethard, Steven  and
      Cotterell, Ryan  and
      Chakraborty, Tanmoy  and
      Zhou, Yichao",
    booktitle = "Proceedings of the 2021 Conference of the North American Chapter of the Association for Computational Linguistics: Human Language Technologies",
    month = jun,
    year = "2021",
    address = "Online",
    publisher = "Association for Computational Linguistics",
    url = "https://aclanthology.org/2021.naacl-main.179/",
    doi = "10.18653/v1/2021.naacl-main.179",
    pages = "2219--2263"
}

@inproceedings{islam-2023-climate,
    author = {Islam, Tunazzina and Zhang, Ruqi and Goldwasser, Dan},
    title = {Analysis of Climate Campaigns on Social Media using Bayesian Model Averaging},
    year = {2023},
    isbn = {9798400702310},
    publisher = {Association for Computing Machinery},
    address = {New York, NY, USA},
    url = {https://doi.org/10.1145/3600211.3604665},
    doi = {10.1145/3600211.3604665},
    booktitle = {Proceedings of the 2023 AAAI/ACM Conference on AI, Ethics, and Society},
    pages = {15–25},
    numpages = {11},
    location = {Montr\'{e}al, QC, Canada},
    series = {AIES '23}
}
\bibliographystyle{acl_natbib}

\appendix

\section{Relational Approach Operational Details}
\label{app:drail_details}
Table \ref{tab:ops} details all operations available to participants using the \citet{pacheco-2023-interactive} system.

\begin{table}[t]
    \centering
    \begin{subtable}{\columnwidth}
     \scalebox{0.60}{\begin{tabular}{|>{\arraybackslash}m{2cm}|>{\arraybackslash}m{9cm}|}
     \hline
       \textbf{Operations}  &  \textbf{Description} \\
       \hline
       Finding Partitions  &  Experts can find partitions in the space of unassigned instances. We currently support the K-means~\cite{Jin2010} and Hierarchical Density-Based Clustering~\cite{mendelsohn-etal-2021-modeling} algorithms. 
       \\
       \hline
       Text-based Queries & Experts can type any query in natural language and find instances that are close to the query in the embedding space. \\
       \hline
       Finding Similar Instances & Experts have the ability to select each instance and find other examples that are close in the embedding space.\\      
       \hline
       Listing Themes and Instances  & Experts can browse the current list of themes and their mapped instances. Instances are ranked in order of ``goodness'', corresponding to the similarity in the embedding space to the theme representation. They can be listed from closest to most distant, or from most distant to closest.  \\
       \hline
       Visualizing Local Explanations & Experts can visualize aggregated statistics and explanations for each of the themes. To obtain these explanations, we aggregate all instances that have been identified as being associated with a theme. Explanations include wordclouds, frequent entities and their sentiments, and graphs of concept distributions. \\
       \hline
       Visualizing Global Explanations & Experts can visualize aggregated statistics and explanations for the global state of the system. To do this, we aggregate all instances in the database. Explanations include theme distribution, coverage statistics, and t-sne plots \cite{JMLR:v9:vandermaaten08a}. \\
       \hline
       
    \end{tabular}}
    \caption{Exploratory Operations}\label{tab:exploratory_ops}
    \end{subtable}
    \begin{subtable}{\columnwidth}
\scalebox{0.60}{\begin{tabular}{|>{\arraybackslash}m{2cm}|>{\arraybackslash}m{9cm}|}
     \hline
       \textbf{Operations}  &  \textbf{Description} \\
       \hline
       Adding, Editing and Removing Themes & Experts can create, edit, and remove themes. The only requirement for creating a new theme is to give it a unique name. Similarly, themes can be edited or removed at any point. If any instances are assigned to a theme being removed, they will be moved to the space of unassigned instances.  \\
       \hline
       Adding and Removing Examples  & Experts can assign ``good'' and ``bad'' examples to existing themes. Good examples are instances that characterize the named theme. Bad examples are instances that could have similar wording to a good example, but that have different meaning. Experts can add examples in two ways: they can mark mapped instances as ``good'' or ``bad'', or they can directly contribute example phrases. \\
       \hline
       Adding or Correcting Concepts & We allow users to upload additional observed or predicted concepts for each textual instance. For instances and phrases added as ``good'' and ``bad'' examples, we allow users to add or edit the values of these concepts. The intuition behind this operation is to collect additional information for learning to map instances to themes. \\
       \hline
    \end{tabular}}
    \caption{Intervention Operations}\label{tab:intervention_ops}
    \end{subtable}
    \caption{Interactive Operations for the \citet{pacheco-2023-interactive} System}
    \label{tab:ops}
\end{table}

\section{LLM-based Experimental Configuration and Prompt Details}\label{app:llm_details}


We used the Llama 3.2 3B-Instruct \citep{grattafiori2024llama3herdmodels} model with a batch size of 32 for all generation tasks. In total, we ran 10 jobs at an average compute time of 24 hours per job using an A100 40GB VRAM GPU.

\begin{table}[H]
    \centering
    \scriptsize
    \begin{tabular}{l|l|l}
    \toprule
        \multicolumn{3}{c}{\textit{COVID dataset}} \\
        \midrule
        code-book & Gwet's $AC_1$ & \# Unlabeled Docs\\
        \midrule
        Sync 1   & $0.42$ & $5,548\ (6.5\%)$ \\
        Sync 2   & $0.49$ & $9,766\ (11.4\%)$ \\
        Async 1  & $0.61$ & $611\ (0.7\%)$ \\
        Async 2  & $0.62$ & $2,506\ (2.9\%)$ \\
        Async 3  & $0.46$ & $13,816\ (16.1\%)$ \\
        \midrule
        \multicolumn{3}{c}{\textit{Climate dataset}} \\
        \midrule
        code-book & Gwet's $AC_1$ & \# Unlabeled Docs\\
        \midrule
        Sync 1   & $0.38$ & $291\ (5.3\%)$ \\
        Sync 2   & $0.33$ & $372\ (6.8\%)$ \\
        Async 1  & $0.38$ & $150\ (2.7\%)$ \\
        Async 2  & $0.39$ & $277\ (5.1\%)$ \\
        Async 3  & $0.25$ & $120\ (2.2\%)$ \\
    \bottomrule
    \end{tabular}
    \caption{Results for the selected prompt for each coding session using the LLM-based system. Gwet's $AC_1$ is used to select the best prompt for running the full dataset. The number of unlabeled documents represent documents where the LLM produced a label not created by human annotators after running the full dataset (percentage of the dataset unlabeled).}
    \label{tab:laca-ac1-combined}
\end{table}

\subsection{LLM Prompts}
To choose the best prompt for labeling the full dataset, we run preliminary experiments with each code-book to calculate human-model annotator agreement. The model is prompted to label all of the documents already labeled by the human and agreement is calculated using Gwet's $AC_1$, as shown in Table \ref{tab:laca-ac1-combined}. Figure \ref{fig:llm-prompt-1} is the prompt template adapted from \citet{chew2023llmassistedcontentanalysisusing} while figures \ref{fig:llm-prompt-2}, \ref{fig:llm-prompt-3}, and \ref{fig:llm-prompt-4} are additional templates created based on  the first template.

\begin{figure}[t]
\begin{tcolorbox}[
    colback=gray!5,
    colframe=black,
    boxrule=0.4pt,
    left=4pt, right=4pt, top=4pt, bottom=4pt,
    fontupper=\ttfamily\small,
]
To code this tweet, do the following:\\
- First, read the codebook and the tweet.\\
- Next, decide which code is most applicable and explain your reasoning for the coding decision.\\
- Finally, generate json with your code and your reason for the coding decision. The response MUST be formatted as JSON.

Codes:\\
--\\
<codes>\\
--\\
Codebook:\\
--\\
<codebook>\\
--\\
Tweet:\\
--\\
<tweet>\\
--\\
JSON Output:\\
--\\
"code" : "",\\
"reason" : ""\\
--
\end{tcolorbox}
\caption{Prompt Template 1 for LLM-based experiments.}
\label{fig:llm-prompt-1}
\end{figure}

\begin{figure}[t]
\begin{tcolorbox}[
    colback=gray!5,
    colframe=black,
    boxrule=0.4pt,
    left=4pt, right=4pt, top=4pt, bottom=4pt,
    fontupper=\ttfamily\small,
]
To code this tweet, do the following:\\
\\
First, read the codebook and the tweet. Next, decide which code is most applicable based on the tweet's content and explain your reasoning for the coding decision. Finally, generate a JSON object with the selected code and provide a brief explanation for your coding decision.\\
The response MUST be formatted as JSON.\\
Codebook: \\
Themes: <"theme": "definition">\\
Tweet: < "text": "<text>" >\\
JSON Output: < "code": "", "reason": "" >\\
\end{tcolorbox}
\caption{Prompt Template 2 for LLM-based experiments.}
\label{fig:llm-prompt-2}
\end{figure}

\begin{figure}[t]
\begin{tcolorbox}[
    colback=gray!5,
    colframe=black,
    boxrule=0.4pt,
    left=4pt, right=4pt, top=4pt, bottom=4pt,
    fontupper=\ttfamily\small,
]
To generate code for this tweet, provide a step-by-step explanation of how to approach the task.\\
\\
First, analyze the tweet's content and identify key concepts, such as the type of object or class being described, any specific behaviors or requirements, and relevant keywords.\\
Next, evaluate the codebook options and determine which one is most applicable. Explain your reasoning for your decision, including any similarities between the tweet and the code definitions, or any specific requirements mentioned in the tweet that align with a particular code.\\
Finally, generate a JSON object with the selected code and provide additional context, including:\\
- A clear explanation of how you arrived at your chosen code\\
- Any relevant notes or comments about the code's functionality and requirements\\
- A brief comparison to other codes in the book, if applicable\\
The response MUST be formatted as JSON.\\
Codebook: <codebook>\\
Tweet: <tweet>\\
JSON Output: < "code": "", "reasoning": "", "context": "" >\\
\end{tcolorbox}
\caption{Prompt Template 3 for LLM-based experiments.}
\label{fig:llm-prompt-3}
\end{figure}

\begin{figure}[t]
\begin{tcolorbox}[
    colback=gray!5,
    colframe=black,
    boxrule=0.4pt,
    left=4pt, right=4pt, top=4pt, bottom=4pt,
    fontupper=\ttfamily\small,
]
To analyze this tweet and select a relevant theme, follow these steps:\\
\\
First, read the tweet and identify key concepts, such as emotions, objects, or ideas mentioned in the text.\\
Next, evaluate the theme options and determine which one is most applicable. Explain your reasoning for your decision, including any connections you see between the tweet's content and the theme definitions.\\
Then, generate a JSON object with the selected theme and provide additional insight into your analysis. Include:\\
A clear explanation of how you arrived at your chosen theme\\
Any specific characteristics or keywords from the tweet that support your decision\\
A brief comparison to other themes, if applicable\\
The response MUST be formatted as JSON.\\
Themes: <Codebook>\\
Tweet: <tweet>\\
JSON Output: <"theme": "", "insight": "">\\
\end{tcolorbox}
\caption{Prompt Template 4 for LLM-based experiments.}
\label{fig:llm-prompt-4}
\end{figure}

\section{Dataset Analysis}
\label{app:dataset}

The two corpora exhibit substantial structural differences as seen in Table \ref{tab:statistics}. The climate dataset comprises 5,471 documents with a mean length of 53.1 words, while the COVID dataset contains 85,799 documents averaging 27.5 words. This approximately two-fold difference in document length is statistically significant (Mann-Whitney U, p<0.001, effect size r=-0.353). Climate advertisements also demonstrate greater variability in length (CV=1.47) compared to the more constrained COVID tweets (CV=0.50). Lexical diversity metrics reveal nuanced differences. While type-token ratios are comparable (0.060 vs. 0.051), the Measure of Textual Lexical Diversity (MTLD) indicates higher lexical sophistication in COVID tweets (119.6 vs. 95.6). 

\begin{table}[t]
\resizebox{\columnwidth}{!}{%
\begin{tabular}{@{}l|l|l@{}}
\toprule
\textbf{Metric}                                                                                       & \textbf{Covid} & \textbf{Climate} \\ \midrule
Number of Documents                                                                                   & 85,799                 & 5,471                 \\
Words per Document                                                                                    & 27.50                    & 53.10                   \\
Sentences per Document                                                                                & 2.50                   & 3.40                  \\
Type-Token Ratio                                                                                      & 0.051                 & 0.060                \\
\begin{tabular}[c]{@{}l@{}}Measure of Textual \\ Lexical Diversity \\ (threshold = 0.72)\end{tabular} & 119.65                & 95.58                \\ \bottomrule
\end{tabular}%
}
\caption{Corpus-level statistics for the Covid and Climate datasets.}
\label{tab:statistics}
\end{table}

Readability analyses indicate that climate advertisements present greater cognitive demands. The Flesch Reading Ease score for climate ads (M=47.4) falls within the ``difficult'' range, whereas COVID tweets score higher (M=54.9), indicating moderately easier comprehension (p<0.001, r=0.211). Correspondingly, climate ads require higher grade-level reading ability (Flesch-Kincaid Grade: 10.7 vs. 9.2; Gunning Fog Index: 13.2 vs. 11.2). See Table \ref{tab:readability}.

\begin{table}[t]
\resizebox{\columnwidth}{!}{%
\begin{tabular}{@{}l|l|l@{}}
\toprule
\textbf{Metric}             & \textbf{Covid} & \textbf{Climate} \\ \midrule
Flesch Reading Ease Score   & 54.85                 & 47.38                \\
Flesch Kincaid Grade        & 9.18                  & 10.68                \\
Gunning Fog Index           & 11.20                 & 13.22                \\
Automated Readability Index & 12.34                 & 11.85                \\ \bottomrule
\end{tabular}%
}
\caption{Measures of Linguistic Complexity of the Covid and Climate datasets.}
\label{tab:readability}
\end{table}

We also performed an analysis over a sample of 5000 documents from both datasets to identify the extent of formulaic or template centric content. DBSCAN clustering on TF-IDF similarity matrices (threshold=0.8) identified 137 template clusters in climate ads, encompassing 20.1\% of documents, compared to only 31 clusters (3.7\%) in COVID tweets. The proportion of high-similarity document pairs (>=0.8 cosine similarity) is an order of magnitude greater in climate ads (0.059\% vs. 0.006\%). See Table \ref{tab:template}.

\begin{table}[t]
\resizebox{\columnwidth}{!}{%
\begin{tabular}{@{}l|l|l@{}}
\toprule
\textbf{Metric}                                                                                & \textbf{Covid} & \textbf{Climate} \\ \midrule
Template Clusters                                                                              & 31                    & 137                  \\
\begin{tabular}[c]{@{}l@{}}Percentage of Documents \\ in Template Clusters\end{tabular}        & 3.70                  & 20.10                \\
\begin{tabular}[c]{@{}l@{}}Percentage of High Similarity\\ Document Pairs\end{tabular} & 0.006                  & 0.059                \\
Average Cluster Size                                                                           & 5.96                  & 7.34                 \\
Largest Cluster Size                                                                           & 31                    & 75                   \\ \bottomrule
\end{tabular}%
}
\caption{Template patterns across Covid and Climate datasets.}
\label{tab:template}
\end{table}

Characteristic n-grams reflect domain-specific discourse patterns. Climate ads prominently feature phrases such as ``oil natural gas'', ``clean energy jobs'', and ``fight climate change'', while COVID tweets center on vaccine-related language (``covid 19 vaccine'', ``getting covid vaccine''). These patterns suggest that climate advertising employs standardized messaging templates, whereas COVID discourse largely centers around vaccines. See Table \ref{tab:ngrams}.

\begin{table}[t]
\resizebox{\columnwidth}{!}{%
\begin{tabular}{@{}l|l@{}}
\toprule
\textbf{Covid}                                                                                                                                               & \textbf{Climate}                                                                                                                                      \\ \midrule
\textit{\begin{tabular}[c]{@{}l@{}}'covid 19 vaccine', 'getting covid vaccine', \\ 'covid vaccine https', 'got covid vaccine', \\ 'covid 19 vaccines'\end{tabular}} & \textit{\begin{tabular}[c]{@{}l@{}}'https bit ly', 'oil natural gas', \\ 'clean energy jobs', 'fight climate change', \\ 'oil gas industry'\end{tabular}} \\ \bottomrule
\end{tabular}%
}
\caption{Top n-gram phrases in Covid and Climate datasets.}
\label{tab:ngrams}
\end{table}

Gaussian Mixture Model analysis with SBERT embeddings \citep{reimers-gurevych-2019-sentence} reveals that both corpora exhibit high thematic clarity, with mean maximum cluster probabilities exceeding 0.98. However, COVID tweets demonstrate marginally higher semantic entropy (M=0.042 vs. 0.023), indicating slightly greater theme ambiguity. The proportion of documents spanning multiple clusters (probability difference < 0.2 between top-2 clusters) is higher for COVID tweets (0.66\% vs. 0.28\%). Silhouette scores are low for both corpora (0.022 vs. 0.018), suggesting that while documents cluster clearly into dominant themes, inter-cluster boundaries are not sharply defined. See Table \ref{tab:theme-ambiguity}.

\begin{table}[t]
\resizebox{\columnwidth}{!}{%
\begin{tabular}{@{}l|l|l@{}}
\toprule
\textbf{Metric}                                                                                                       & \textbf{Covid} & \textbf{Climate} \\ \midrule
\begin{tabular}[c]{@{}l@{}}Average Entropy across \\ Documents\end{tabular}                                           & 0.042      & 0.023                \\
\begin{tabular}[c]{@{}l@{}}Average Confidence in \\ Primary Cluster Assignment\end{tabular}                           & 0.989                 & 0.994                \\
\begin{tabular}[c]{@{}l@{}}Percentage of Documents \\ Where Top 2 Clusters are \\ within 0.2 Probability\end{tabular} & 0.660                 & 0.280                \\
Silhouette Score                                                                                                      & 0.022                 & 0.018                \\ \bottomrule
\end{tabular}%
}
\caption{Semantic theme ambiguity for Covid and Climate datasets.}
\label{tab:theme-ambiguity}
\end{table}

Mean pairwise cosine similarity in the SBERT embedding space is substantially higher for COVID tweets (0.429 vs. 0.261), indicating greater semantic homogeneity. The inter-quartile range confirms this pattern: COVID tweets exhibit similarity scores between 0.365 -- 0.518, while climate ads range from 0.167 -- 0.348. Notably, 62.6\% of climate ad pairs fall below 0.3 similarity, compared to only 14.1\% of COVID tweet pairs. Document-to-centroid similarity further corroborates this finding (0.654 vs. 0.510), demonstrating that COVID tweets cluster more tightly around their corpus centroid. See Table \ref{tab:similarity-dist}.

\begin{table}[H]
\resizebox{\columnwidth}{!}{%
\begin{tabular}{@{}l|l|l@{}}
\toprule
\textbf{Metric}                                                                             & \textbf{Covid} & \textbf{Climate} \\ \midrule
\begin{tabular}[c]{@{}l@{}}Average Pairwise Similarity\\ between documents\end{tabular}     & 0.43                  & 0.26                 \\
\begin{tabular}[c]{@{}l@{}}25th Percentile Pairwise \\ Similarity Score\end{tabular}        & 0.37                  & 0.17                 \\
\begin{tabular}[c]{@{}l@{}}75th Percentile Pairwise \\ Similarity Score\end{tabular}        & 0.52                  & 0.35                 \\
\begin{tabular}[c]{@{}l@{}}Average Similarity of\\ documents with the centroid\end{tabular} & 0.65                  & 0.51                 \\
\begin{tabular}[c]{@{}l@{}}Percentage of Pairs with\\ Similarity > 0.7\end{tabular}         & 0.70                  & 0.20                 \\
\begin{tabular}[c]{@{}l@{}}Percentage of Pairs with\\ Similarity > 0.3\end{tabular}         & 14.14                 & 62.55                \\ \bottomrule
\end{tabular}%
}
\caption{Semantic Similarity Distribution for Covid and Climate datasets.}
\label{tab:similarity-dist}
\end{table}


\section{Themes by Code-book}
\label{app:theme-counts}

Tables~\ref{tab:codebook-covid-tm} through \ref{tab:codebook-climate-llm} list the final themes from every experiment, organized by dataset and model. Each table covers one (dataset, model) combination and reports themes across the three asynchronous and two synchronous sessions. Theme counts are summarized in Table~\ref{tab:theme-counts}, and themes are reported verbatim as coders named them.

%
%

\begin{table*}[p]
\centering
\scriptsize
\renewcommand{\arraystretch}{1.1}
\setlength{\tabcolsep}{4pt}
\begin{tabularx}{\textwidth}{@{}YYYYY@{}}
\toprule
\textbf{Async 1} & \textbf{Async 2} & \textbf{Async 3} & \textbf{Sync 1} & \textbf{Sync 2} \\
\midrule
1.~Pro Trump and against Biden/Obama \newline 2.~Dispelling misinformation regarding death \newline 3.~Vaccine availability \newline 4.~2nd dose side effects \newline 5.~Comparisons to other virus vaccines \newline 6.~Anger over mask mandates \newline 7.~Reminders of the 2nd vaccine \newline 8.~Origin of the vaccine \newline 9.~Discussing vaccines in schools \newline 10.~Positive outcomes of vaccine \newline 11.~Concerns regarding Johnson and Johnson vaccine \newline 12.~CVS job openings \newline 13.~Information aimed towards minority communities \newline 14.~Against anti-vaxxers \newline 15.~Calling out liars \newline 16.~Free food for vaccinated people \newline 17.~Vaccine card mandates \newline 18.~Vaccine hesistancy \newline 19.~Vaccine appointment info & 1.~VaccineRolloutNews \newline 2.~AntiTrump \newline 3.~VaccineSideEffects \newline 4.~VaccineTestScience \newline 5.~HopefulPeopleGetVaccinated \newline 6.~WhereToGetVaccine \newline 7.~FDANews\&Discourse \newline 8.~VaccinePassportNews \newline 9.~FakeNews \newline 10.~VaccineAdvocacy \newline 11.~AntiRepublican \newline 12.~AntiDemocrat \newline 13.~PandemicNews \newline 14.~VaccineHesitancy \newline 15.~POCVaccineAwareness \newline 16.~FightingDisinformation \newline 17.~VaccineChildRisk \newline 18.~VaccineProblemsNews & 1.~Anti-Biden and Pro-Trump Sentiment \newline 2.~Comparing the COVID Vaccine to other Deadly Disease Vaccines (polio, measles) \newline 3.~Anger towards COVID Deniers / COVID Deniers Passing Away \newline 4.~Immediate After-Effects from COVID Vaccine (soreness, nausea) \newline 5.~Family Members Receiving the COVID Vaccine \newline 6.~Vaccine Eligibility Updates \newline 7.~Proof of Vaccination (vaccine card/passport) \newline 8.~Adverse Childhood Experience's being caused by COVID \newline 9.~COVID Vaccine causing Blood Clots \newline 10.~Unknown Long-Term Effects of COVID Vaccine \newline 11.~Become a Pharmacy Technician Alerts \newline 12.~COVID FDA Approval Status \newline 13.~Anger Towards Republicans Lying about COVID \newline 14.~Giving Info about COVID and the Vaccine to Minorities in America \newline 15.~Walgreens Vaccine Appointment Updates & 1.~trump supporters saying biden is wrongfully taking credit for vaccine \newline 2.~side effects of vaccine \newline 3.~announcing vaccine eligibility \newline 4.~wholesome reactions to getting the vaccine \newline 5.~inefficacy of vaccine \newline 6.~corelation between trauma and covid \newline 7.~halting vaccination due to adverse effects \newline 8.~proof of vaccine \newline 9.~appeal to get vaccinated \newline 10.~covid related deaths \newline 11.~advertising for covid related jobs \newline 12.~freebies with proof of vaccine \newline 13.~concerns about effects of vaccines \newline 14.~disproportionate impact and equitable recover across racial lines \newline 15.~political conspiracy / nicki minaj controversy \newline 16.~vaccine appointments & 1.~criticism of governments' response to vaccine \newline 2.~family covid vaccine experience \newline 3.~immediate personal side effects \newline 4.~state vaccine eligibility \newline 5.~non-covid vaccine comparision \newline 6.~institutional adverse childhood experiences \newline 7.~vaccine certification \newline 8.~anti-vaccine conspiracy outcomes \newline 9.~pharmacy hiring \newline 10.~FDA vaccine approval \newline 11.~long term side effects \newline 12.~community vaccine equity \newline 13.~vaccine side effects / Nicki Minaj \newline 14.~covid appointment \\
\bottomrule
\end{tabularx}
\caption{Final codebooks for the COVID Tweets dataset using the Topic Model approach.}
\label{tab:codebook-covid-tm}
\end{table*}

\begin{table*}[p]
\centering
\scriptsize
\renewcommand{\arraystretch}{1.1}
\setlength{\tabcolsep}{4pt}
\begin{tabularx}{\textwidth}{@{}YYYYY@{}}
\toprule
\textbf{Async 1} & \textbf{Async 2} & \textbf{Async 3} & \textbf{Sync 1} & \textbf{Sync 2} \\
\midrule
1.~Receiving first dose of covid vaccine \newline 2.~Criticising political figures \newline 3.~Promoting covid related news articles \newline 4.~Vaccine accessibility and distribution \newline 5.~Information about covid vaccine availability and appointments \newline 6.~Encouraging people to get the covid vaccine \newline 7.~The vaccine does not work because you can still get covid \newline 8.~Skepticism over the covid vaccine \newline 9.~Experiences of vaccine side effects \newline 10.~Praising frontline healthcare workers \newline 11.~Saying that you can still get covid if you have the vaccine \newline 12.~Receiving second vaccine dose \newline 13.~Encouraging getting the vaccine \newline 14.~Criticising The President \newline 15.~FDA Approval of covid vaccine \newline 16.~Providing resources to those who are not involved in the covid debate \newline 17.~Praising Government Leadership \newline 18.~Against covid disbelievers \newline 19.~Stating that the vaccine does not cause covid & 1.~GotVaccinated \newline 2.~AntiRepublican \newline 3.~WhereToGetVaccine \newline 4.~AdvocateForVaccine \newline 5.~VaccineSymptomReport \newline 6.~VaccineEfficacyDenial \newline 7.~VaccineRefusal \newline 8.~VaccineKills \newline 9.~AntiDemocrat \newline 10.~BlameFauci & 1.~ReceivedFirstDoseOfCovidVaccine \newline 2.~RepublicansDownplayingCovid \newline 3.~FdaApprovesCovidVaccine \newline 4.~GettingTheCovidVaccine \newline 5.~NewsArticlesAboutCovidVaccineProgress \newline 6.~ProCovidVaccine \newline 7.~MentionsAMayor \newline 8.~NewsAgenciesReportingOnTheCovidVaccine \newline 9.~AntiCovidVaccine \newline 10.~CovidVaccineAftereffects \newline 11.~GetYourCovidVaccine \newline 12.~BidenAdministrationEnforcingCovidVaccine \newline 13.~CovidVaccineOnlyReducesTheEffectsOfCovid \newline 14.~HealthcareWorkers \newline 15.~AntiBiden \newline 16.~NegativeViewOnTrump \newline 17.~RepublicansResponsibleForCovidDeaths \newline 18.~IsraelOutbreak & 1.~VaxSymptoms \newline 2.~GovGoodPolicies \newline 3.~VaxAppointmentInfo \newline 4.~VaxApprovalInfo \newline 5.~VaxDoesntWork \newline 6.~UnjustifiedFearOfVax \newline 7.~IGotTheVax \newline 8.~GovBadPolicies \newline 9.~VaxLessensSymptoms & 1.~PostVaxSymptoms \newline 2.~ReasonsForUSLaggingOnVaccines \newline 3.~VaxDistributionIssueDueToLocalPolicy \newline 4.~VaxAvailabilityInfo \newline 5.~\#IGotMyVaccine \newline 6.~VaxDoesMoreHarmThanGood \newline 7.~VaxLessensSymptoms \newline 8.~FDAapproval \\
\bottomrule
\end{tabularx}
\caption{Final codebooks for the COVID Tweets dataset using the Relational approach.}
\label{tab:codebook-covid-rel}
\end{table*}

\begin{table*}[p]
\centering
\scriptsize
\renewcommand{\arraystretch}{1.1}
\setlength{\tabcolsep}{4pt}
\begin{tabularx}{\textwidth}{@{}YYYYY@{}}
\toprule
\textbf{Async 1} & \textbf{Async 2} & \textbf{Async 3} & \textbf{Sync 1} & \textbf{Sync 2} \\
\midrule
1.~Receiving first dose of covid vaccine \newline 2.~Praising Government Leadership \newline 3.~Providing resources to those who are not involved in the covid debate \newline 4.~Vaccine accessibility and distribution \newline 5.~Praising frontline healthcare workers \newline 6.~Promoting covid related news articles \newline 7.~Information about covid vaccine availability and appointments \newline 8.~Stating that the vaccine does not cause covid \newline 9.~Skepticism over the covid vaccine \newline 10.~Criticising The President \newline 11.~Experiences of vaccine side effects \newline 12.~Receiving second vaccine dose \newline 13.~Encouraging getting the vaccine \newline 14.~Saying that you can still get covid if you have the vaccine \newline 15.~The vaccine does not work because you can still get covid & 1.~GotVaccinated \newline 2.~BlameFauci \newline 3.~VaccineRefusal \newline 4.~WhereToGetVaccine \newline 5.~VaccineEfficacyDenial \newline 6.~AdvocateForVaccine \newline 7.~VaccineSymptomReport \newline 8.~AntiDemocrat \newline 9.~AntiRepublican \newline 10.~VaccineKills & 1.~received\_vax \newline 2.~negative\_discourse\_around\_politicians \newline 3.~vax\_appointment\_availability \newline 4.~vax\_approved \newline 5.~vax\_reduces\_fatality \newline 6.~positive\_discourse\_around\_politicians \newline 7.~news\_about\_vax \newline 8.~anti\_vax \newline 9.~encourage\_getting\_vax \newline 10.~side\_effects\_of\_vaccine \newline 11.~vax\_is\_ineffective\_or\_harmful \newline 12.~getting\_attention\_of\_politician \newline 13.~sharing\_stories\_related\_to\_covid & 1.~GetAVaccine \newline 2.~CovidLiberal \newline 3.~ScrewYouGovernment \newline 4.~CovidDebunking \newline 5.~GoodNewsAboutVaccines \newline 6.~VaccineAppointmentAvailable \newline 7.~GotFirstDose \newline 8.~JudgingUnvaccinated \newline 9.~ThankYouGovernment \newline 10.~VaccineSymptomsNegative \newline 11.~ThankYouDoctors \newline 12.~CovidConservative \newline 13.~VaccineDoesntPreventCovidNegative \newline 14.~VaccineConspiracy \newline 15.~VaccineSymptomsPositive & 1.~vaccine\_get\_a\_vaccine \newline 2.~vaccine\_politics \newline 3.~vaccine\_rollout \newline 4.~vaccine\_efficacy \newline 5.~vaccine\_first\_dose \newline 6.~vaccine\_symptoms \newline 7.~online\_references \\
\bottomrule
\end{tabularx}
\caption{Final codebooks for the COVID Tweets dataset using the LLM-based approach.}
\label{tab:codebook-covid-llm}
\end{table*}

\begin{table*}[p]
\centering
\scriptsize
\renewcommand{\arraystretch}{1.1}
\setlength{\tabcolsep}{4pt}
\begin{tabularx}{\textwidth}{@{}YYYYY@{}}
\toprule
\textbf{Async 1} & \textbf{Async 2} & \textbf{Async 3} & \textbf{Sync 1} & \textbf{Sync 2} \\
\midrule
1.~AntiGunBan \newline 2.~AntiWoke \newline 3.~ProtectEnvironment \newline 4.~CleanEnergyWorks \newline 5.~ActionAgainstDemSpending \newline 6.~IndigenousAndBlackAdvocacy \newline 7.~PassBidenAct \newline 8.~AntiOilPolitician \newline 9.~AntiPropertyTax \newline 10.~ProUSOilIndustry \newline 11.~ProtectRivers \newline 12.~TPUSAAd \newline 13.~EVCharging \newline 14.~OilTheft & 1.~pro-2A \newline 2.~donations for climate change \newline 3.~solar energy \newline 4.~advocating for city level politics \newline 5.~political activism to oppose climate change \newline 6.~criticism of liberal tax policy \newline 7.~indigenous climate justice \newline 8.~promoting legislation against climate change \newline 9.~pros of wind and solar energy \newline 10.~criticism for disfavoring US ONG production \newline 11.~kudos for confronting ONG emissions \newline 12.~access to water and promoting renewable energy \newline 13.~discussing problems with school board \newline 14.~promoting news and oped program & 1.~AntiGunBan \newline 2.~BidenPolicies \newline 3.~SaveEnvironment \newline 4.~AntiLiberalElectionCandidates \newline 5.~RisingEnergyCost \newline 6.~ProRenewableEnergy \newline 7.~PoliticalLeadershipTraining \newline 8.~BioFuelTheft \newline 9.~ProUSOil \newline 10.~MinorityGroupWomenAdvocacy \newline 11.~DonationsForProRiverPoliticians \newline 12.~SupportForBuildBackBetterAct \newline 13.~WaterPollutionAndContamination \newline 14.~ProCleanEnergyPolitician \newline 15.~TPUSALiveAd \newline 16.~AntiLiberalSpending \newline 17.~AntiPropertyTax \newline 18.~EnergyInfrastructureProjects & 1.~AntiGunBan \newline 2.~EnvironmentalConservation \newline 3.~SchoolBoardFinances/GasInfrastructure \newline 4.~AntiGridDeregulation/PoliticalActionAds \newline 5.~CleanEnergyGoals \newline 6.~BellbrookSugarcreekConservativeElections \newline 7.~Unexplainable \newline 8.~ConservativePoliticsEvents \newline 9.~TPUSAAd \newline 10.~DomesticOilAndGasProduction \newline 11.~ClimateChange \newline 12.~Pipeline \newline 13.~BuildBackBetterAct \newline 14.~AntiGridDeregulation \newline 15.~LiberalTaxHikes \newline 16.~NativeAmericanCulturalPreservation \newline 17.~OilTheft/ClimateLeader \newline 18.~NaturalGasIncentive/CleanEnergyPetition \newline 19.~CleanEnergyProjects \newline 20.~ColoradoAirQuality/SupportingBlackWomen & 1.~gun rights \newline 2.~climate action \newline 3.~renewables \newline 4.~inflation \newline 5.~anti-woke \newline 6.~organized cooking oil theft \newline 7.~indigenous activism \newline 8.~pro oil and natural gas \newline 9.~TPUSA \newline 10.~conservative school board \\
\bottomrule
\end{tabularx}
\caption{Final codebooks for the Climate Ads dataset using the Topic Model approach.}
\label{tab:codebook-climate-tm}
\end{table*}

\begin{table*}[p]
\centering
\scriptsize
\renewcommand{\arraystretch}{1.1}
\setlength{\tabcolsep}{4pt}
\begin{tabularx}{\textwidth}{@{}YYYYY@{}}
\toprule
\textbf{Async 1} & \textbf{Async 2} & \textbf{Async 3} & \textbf{Sync 1} & \textbf{Sync 2} \\
\midrule
1.~AdvocateForGreenEnergy \newline 2.~GreenPowerExpensive \newline 3.~ClimateActionDonation \newline 4.~GreenEnergyWorks \newline 5.~AdvocateForGreenEnergyPoliticalAction \newline 6.~AntiTax \newline 7.~GreenEnergySavesMoney \newline 8.~StopOilPollution \newline 9.~DemocratOutreach \newline 10.~OilCreatesJobs \newline 11.~NuclearAdvocacy \newline 12.~AntiWoke & 1.~AgencyToStopClimateChange \newline 2.~CallsToInvestInRenewableEnergy \newline 3.~NeedQuickResponseForClimateCrisis \newline 4.~PoliticalActivismForRenewableEnergy \newline 5.~DonateToSupportClimateChangeActivism \newline 6.~ChallengesFacingRenewableEnergy \newline 7.~PoliticalActivismForOilAndGas \newline 8.~SupportForRenewableEnergy \newline 9.~ProtectFromHarmsCausedByFossilFuels \newline 10.~InroadsMadeByOilAndGas \newline 11.~InroadsMadeByRenewableEnergy \newline 12.~EconomicAdvantagesOfRenewableEnergy \newline 13.~SupportForBuildBackBetterAct \newline 14.~SupportForOilAndGas \newline 15.~EconomicAdvantagesOfOilAndGas \newline 16.~InroadsMadeByNuclearEnergy \newline 17.~EconomicAdvantagesOfNuclearEnergy & 1.~LegislatorsMustActAgainstClimateChange \newline 2.~ClimateChangeDonations \newline 3.~OilAndGasExtractionRisks \newline 4.~NuclearEnergyBenefits \newline 5.~TackleClimateChange \newline 6.~ReduceEmissions \newline 7.~VirginiaConservativeCampaign \newline 8.~OilAndGasJobs \newline 9.~SouthwestGasInvestment \newline 10.~RenewableEnergyInfrastructureGrowth \newline 11.~NuclearEnergyEducation \newline 12.~OilAndGasTax \newline 13.~BuildBetterAct \newline 14.~RestoreVirginia & 1.~DonationMatchingForClimate \newline 2.~AbandonedWells \newline 3.~GreenEnergyIsCostEffective \newline 4.~ClimateEmergency \newline 5.~NuclearRevenueSupportsJobs \newline 6.~HazardsOfWells \newline 7.~VirginiaConservatives \newline 8.~DemandClimateActionFromLeaders \newline 9.~PublicLandExploitation \newline 10.~GreenEnergyCreatesJobs \newline 11.~OilAndGasSupportJobs \newline 12.~AntiFossilFuelInfrastructure \newline 13.~CleanOilAndGas \newline 14.~GreenEnergyCapacity \newline 15.~CutMethane \newline 16.~CommerceCityDemocrats \newline 17.~RestoreVirginia \newline 18.~ThankRepForBuildBackBetter \newline 19.~BellbrookTaxRevolt \newline 20.~BellbrookWokeRadicals \newline 21.~GreenEnergySupportsCommunities \newline 22.~NuclearEnergyProvidesTaxRevenue \newline 23.~TellRepToNotRaiseOilGasTaxes \newline 24.~OilAndGasProvidesWages \newline 25.~CelebrateBuildBackBetter & 1.~ClimateChangeMovements \newline 2.~ClimateChangeDonations \newline 3.~OilAndGasTaxesBad \newline 4.~LawmakersAndOilAndGas \newline 5.~RenewableEnergyPromotionAndGrowth \newline 6.~CleanEnergyPromotion \newline 7.~RenewableEnergyEconomicBenefits \newline 8.~PoliticalTraining \newline 9.~CallToActionAgainstOilAndGasIndustries \newline 10.~CarbonEmissionReduction \newline 11.~OilAndGasBoostEconomy \newline 12.~PropertyTaxCampaign \newline 13.~BuildBackBetterActCampaign \newline 14.~RightWingCampaigns \newline 15.~NuclearPowerOtherBenefits \newline 16.~NuclearPowerEconomicBenefits \\
\bottomrule
\end{tabularx}
\caption{Final codebooks for the Climate Ads dataset using the Relational approach.}
\label{tab:codebook-climate-rel}
\end{table*}

\begin{table*}[p]
\centering
\scriptsize
\renewcommand{\arraystretch}{1.1}
\setlength{\tabcolsep}{4pt}
\begin{tabularx}{\textwidth}{@{}YYYYY@{}}
\toprule
\textbf{Async 1} & \textbf{Async 2} & \textbf{Async 3} & \textbf{Sync 1} & \textbf{Sync 2} \\
\midrule
1.~AntiWoke \newline 2.~GreenEnergySavesMoney \newline 3.~DemocratOutreach \newline 4.~AdvocateForGreenEnergy \newline 5.~GreenPowerExpensive \newline 6.~GreenEnergyWorks \newline 7.~StopOilPollution \newline 8.~AdvocateForGreenEnergyPoliticalAction \newline 9.~ClimateActionDonation \newline 10.~NuclearAdvocacy \newline 11.~AntiTax \newline 12.~OilCreatesJobs & 1.~AgencyToStopClimateChange \newline 2.~PoliticalActivismForRenewableEnergy \newline 3.~EconomicAdvantagesOfRenewableEnergy \newline 4.~ChallengesFacingRenewableEnergy \newline 5.~InroadsMadeByRenewableEnergy \newline 6.~CallsToInvestInRenewableEnergy \newline 7.~ProtectFromHarmsCausedByFossilFuels \newline 8.~NeedQuickResponseForClimateCrisis \newline 9.~DonateToSupportClimateChangeActivism \newline 10.~EconomicAdvantagesOfOilAndGas \newline 11.~PoliticalActivismForOilAndGas \newline 12.~InroadsMadeByOilAndGas \newline 13.~SupportForBuildBackBetterAct \newline 14.~EconomicAdvantagesOfNuclearEnergy \newline 15.~InroadsMadeByNuclearEnergy \newline 16.~SupportForOilAndGas \newline 17.~SupportForRenewableEnergy & 1.~NuclearEnergyBenefits \newline 2.~BuildBetterAct \newline 3.~ReduceEmissions \newline 4.~NuclearEnergyEducation \newline 5.~TackleClimateChange \newline 6.~OilAndGasExtractionRisks \newline 7.~VirginiaConservativeCampaign \newline 8.~RenewableEnergyInfrastructureGrowth \newline 9.~OilAndGasTax \newline 10.~OilAndGasJobs \newline 11.~SouthwestGasInvestment \newline 12.~ClimateChangeDonations \newline 13.~LegislatorsMustActAgainstClimateChange \newline 14.~RestoreVirginia & 1.~VirginiaConservatives \newline 2.~CommerceCityDemocrats \newline 3.~BellbrookWokeRadicals \newline 4.~BellbrookTaxRevolt \newline 5.~RestoreVirginia \newline 6.~GreenEnergyCreatesJobs \newline 7.~GreenEnergyIsCostEffective \newline 8.~GreenEnergyCapacity \newline 9.~GreenEnergySupportsCommunities \newline 10.~AbandonedWells \newline 11.~AntiFossilFuelInfrastructure \newline 12.~HazardsOfWells \newline 13.~PublicLandExploitation \newline 14.~CutMethane \newline 15.~ClimateEmergency \newline 16.~DemandClimateActionFromLeaders \newline 17.~DonationMatchingForClimate \newline 18.~OilAndGasSupportJobs \newline 19.~OilAndGasProvidesWages \newline 20.~TellRepToNotRaiseOilGasTaxes \newline 21.~CleanOilAndGas \newline 22.~NuclearRevenueSupportsJobs \newline 23.~NuclearEnergyProvidesTaxRevenue \newline 24.~CelebrateBuildBackBetter \newline 25.~ThankRepForBuildBackBetter & 1.~BuildBackBetterActCampaign \newline 2.~CallToActionAgainstOilAndGasIndustries \newline 3.~CarbonEmissionReduction \newline 4.~CleanEnergyPromotion \newline 5.~ClimateChangeDonations \newline 6.~ClimateChangeMovements \newline 7.~LawmakersAndOilAndGas \newline 8.~NuclearPowerEconomicBenefits \newline 9.~NuclearPowerOtherBenefits \newline 10.~OilAndGasBoostEconomy \newline 11.~OilAndGasTaxesBad \newline 12.~PoliticalTraining \newline 13.~PropertyTaxCampaign \newline 14.~RenewableEnergyEconomicBenefits \newline 15.~RenewableEnergyPromotionAndGrowth \newline 16.~RightWingCampaigns \\
\bottomrule
\end{tabularx}
\caption{Final codebooks for the Climate Ads dataset using the LLM-based approach.}
\label{tab:codebook-climate-llm}
\end{table*}

\begin{table*}[t]
\centering
\small
\begin{tabular}{llcccccc}
\toprule
\textbf{Dataset} & \textbf{Model} & \textbf{Async 1} & \textbf{Async 2} & \textbf{Async 3} & \textbf{Sync 1} & \textbf{Sync 2} & \textbf{Total} \\
\midrule
\multirow{3}{*}{COVID Tweets} & Topic Model & 19 & 18 & 15 & 16 & 14 & 82 \\
                              & Relational  & 19 & 10 & 18 &  9 &  8 & 64 \\
                              & LLM-based   & 15 & 10 & 13 & 15 &  7 & 60 \\
\midrule
\multirow{3}{*}{Climate Ads}  & Topic Model & 14 & 14 & 18 & 20 & 10 & 76 \\
                              & Relational  & 12 & 17 & 14 & 25 & 16 & 84 \\
                              & LLM-based   & 12 & 17 & 14 & 25 & 16 & 84 \\
\bottomrule
\end{tabular}
\caption{Number of themes identified per experiment, broken down by dataset, model, and collaboration setting.}
\label{tab:theme-counts}
\end{table*}

\section{Example Heatmaps of Synchronous Code-book Jaccard Similarities}
\label{app:jaccard_examples}
\begin{figure*}[t]
    \centering
    \begin{subfigure}[t]{0.48\textwidth}
        \centering
        \includegraphics[width=\linewidth]{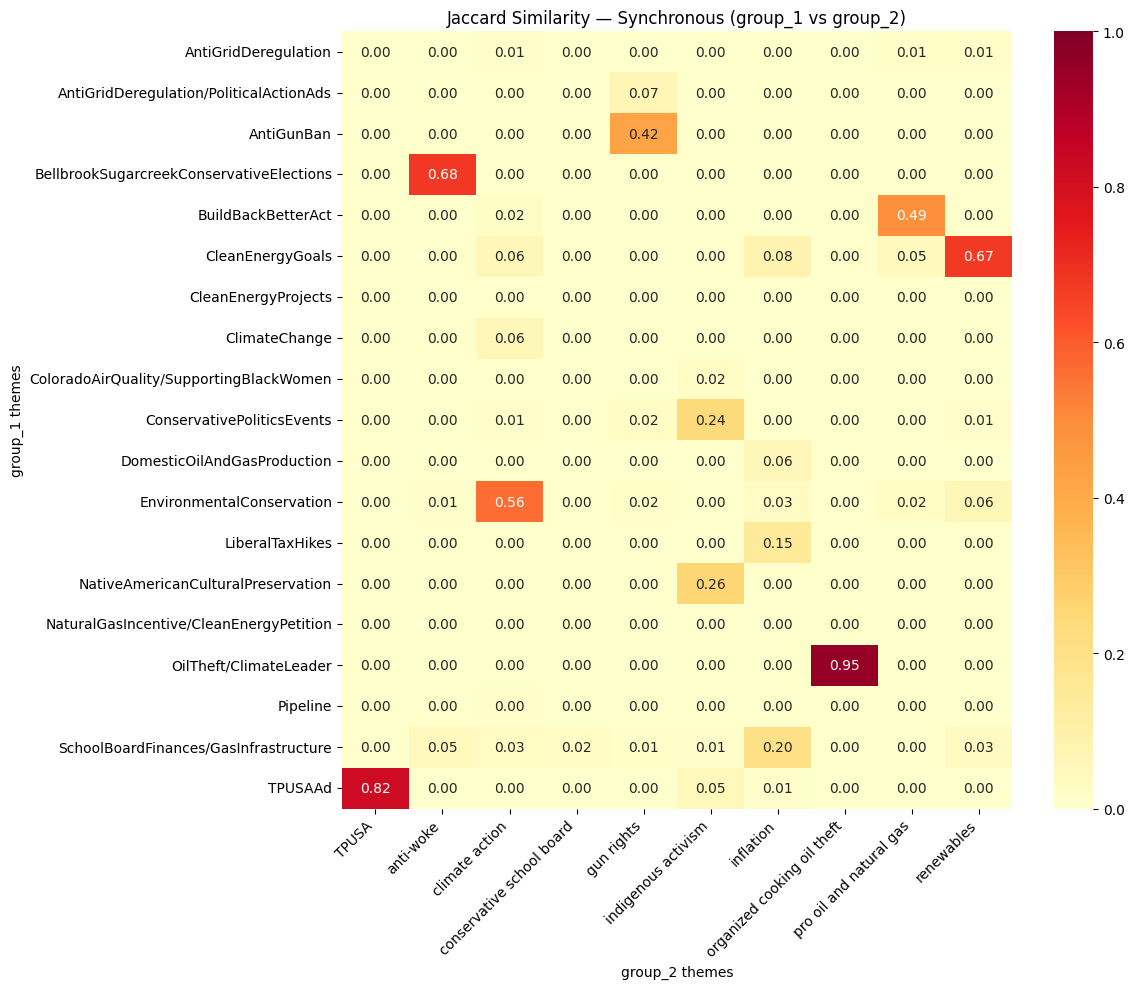}
        \caption{Climate ads.}
        \label{fig:climate_fang_jaccard}
    \end{subfigure}
    \hfill
    \begin{subfigure}[t]{0.48\textwidth}
        \centering
        \includegraphics[width=\linewidth]{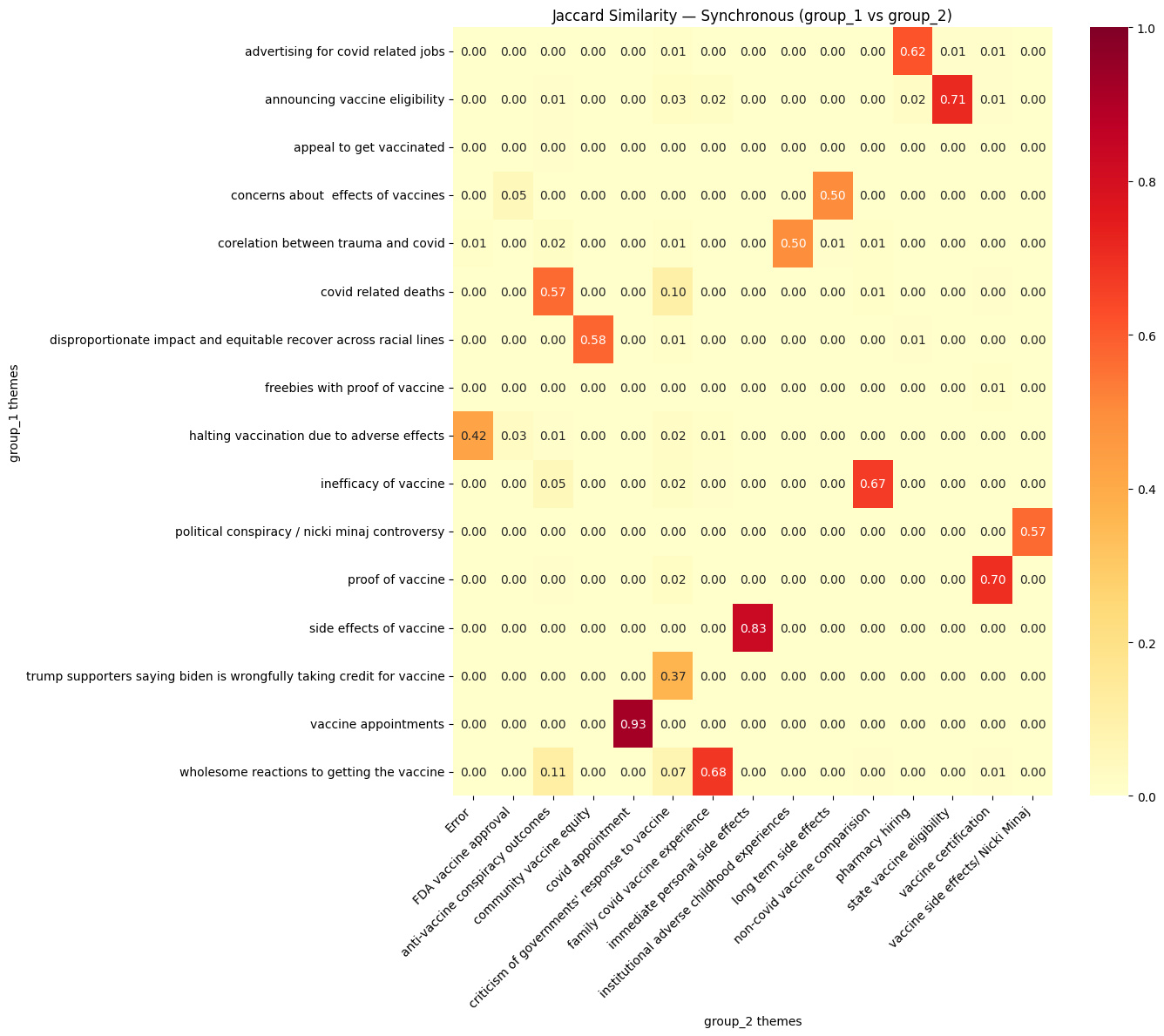}
        \caption{COVID tweets.}
        \label{fig:covid_fang_jaccard}
    \end{subfigure}
    \caption{Jaccard similarity heatmaps using the topic model approach.}
    \label{fig:jaccard_fang}
\end{figure*}

\begin{figure*}[t]
    \centering
    \begin{subfigure}[t]{0.48\textwidth}
        \centering
        \includegraphics[width=\linewidth]{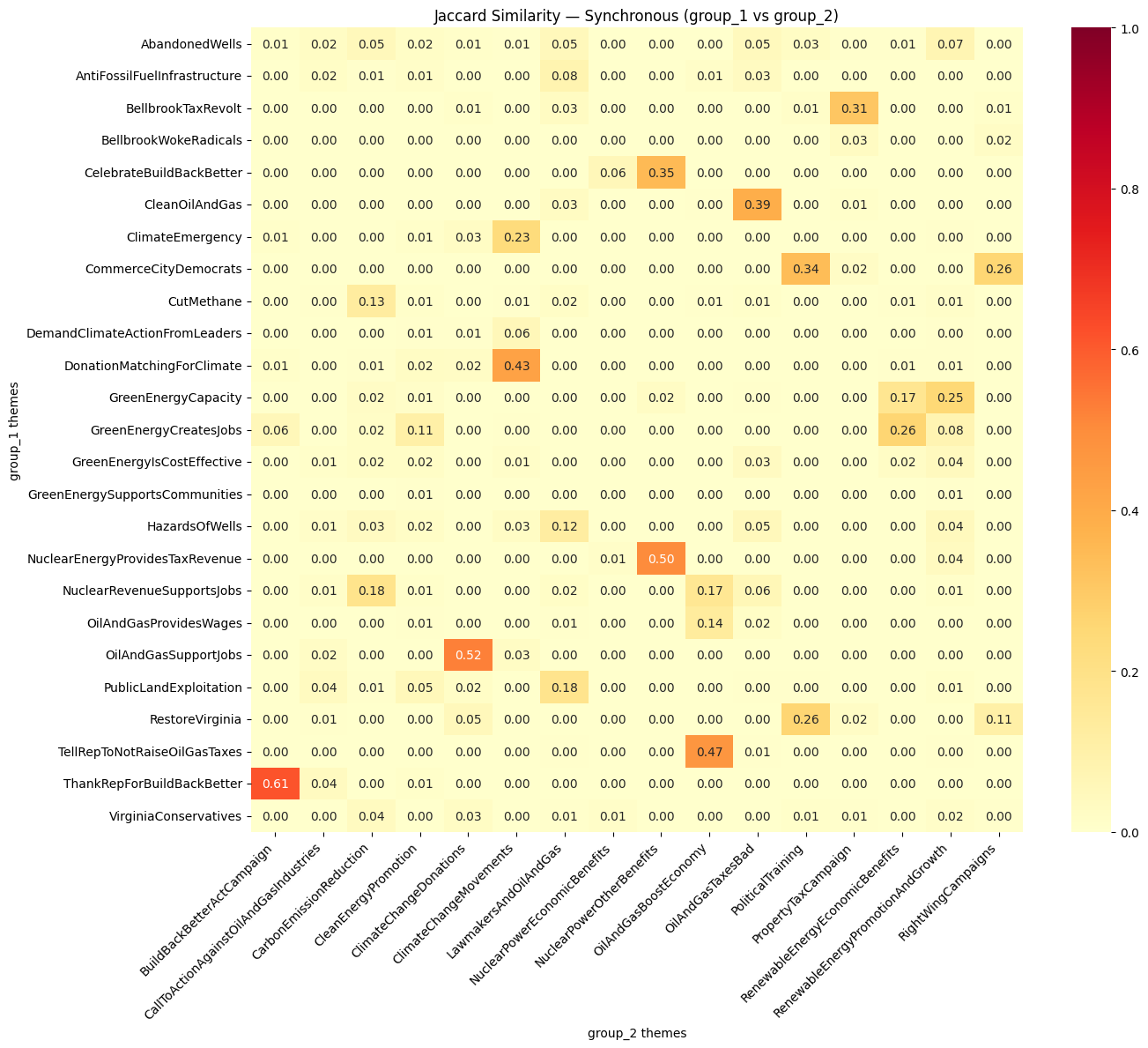}
        \caption{Climate ads.}
        \label{fig:climate_pacheco_jaccard}
    \end{subfigure}
    \hfill
    \begin{subfigure}[t]{0.48\textwidth}
        \centering
        \includegraphics[width=\linewidth]{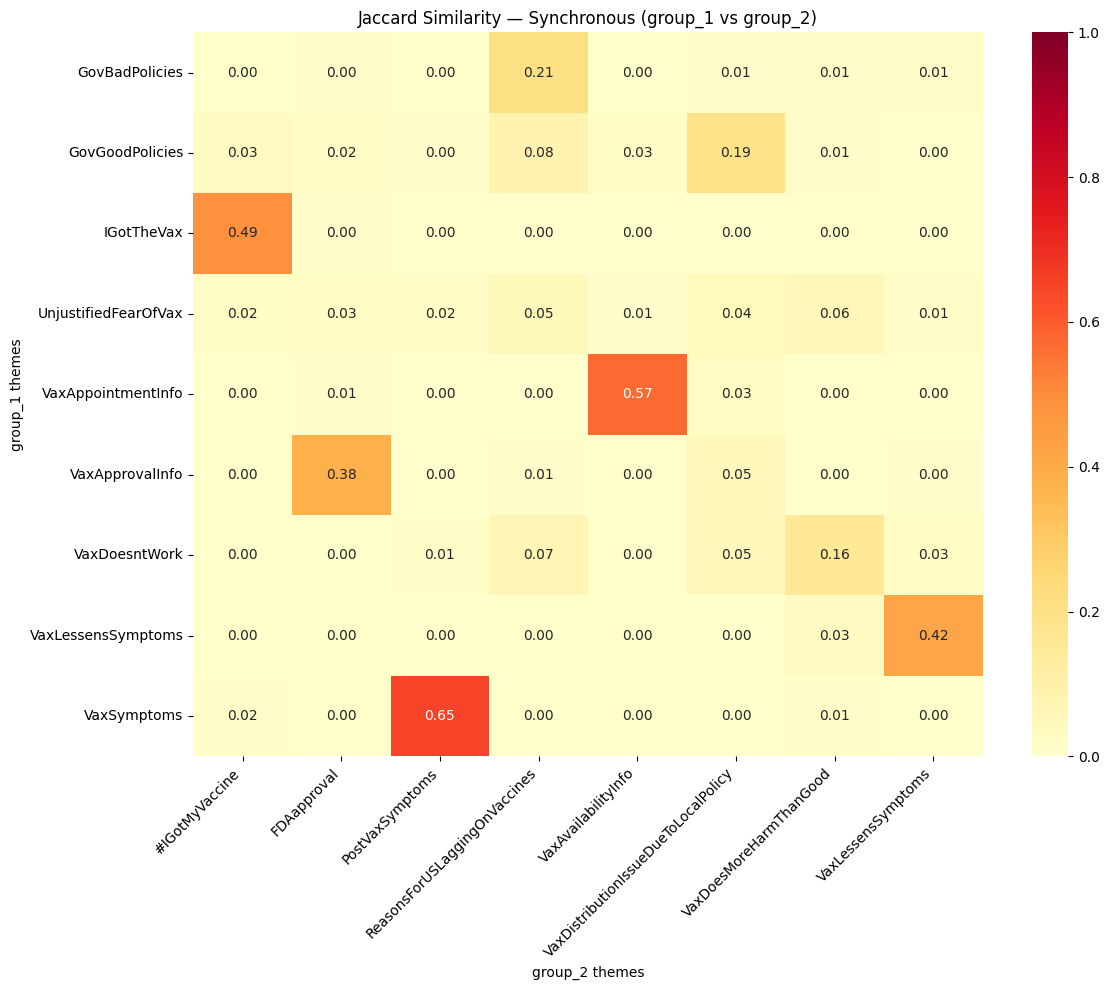}
        \caption{COVID tweets.}
        \label{fig:covid_pacheco_jaccard}
    \end{subfigure}
    \caption{Jaccard similarity heatmaps using the relational approach.}
    \label{fig:jaccard_pacheco}
\end{figure*}

\begin{figure*}[t]
    \centering
    \begin{subfigure}[t]{0.48\textwidth}
        \centering
        \includegraphics[width=\linewidth]{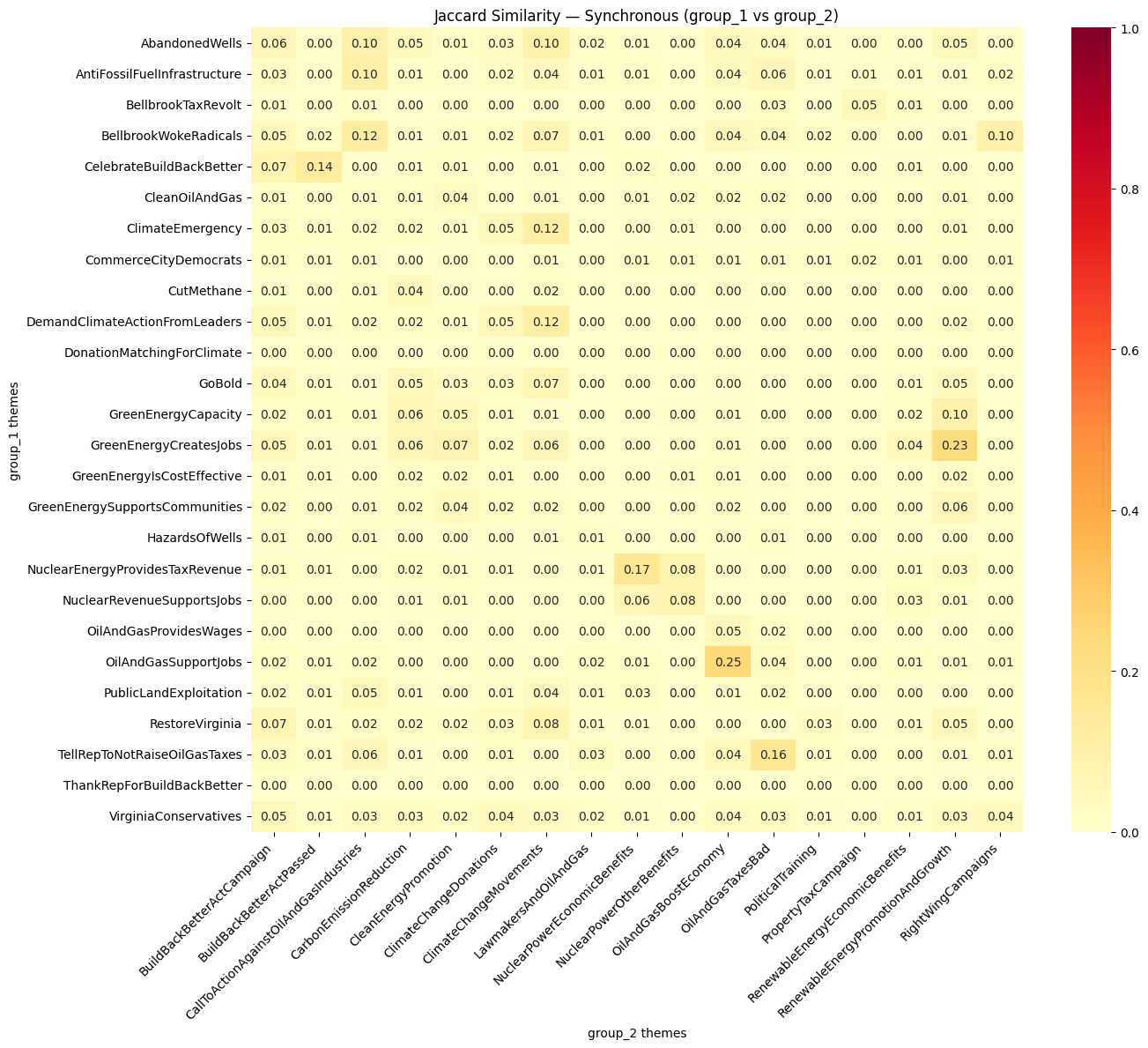}
        \caption{Climate ads.}
        \label{fig:climate_laca_jaccard}
    \end{subfigure}
    \hfill
    \begin{subfigure}[t]{0.48\textwidth}
        \centering
        \includegraphics[width=\linewidth]{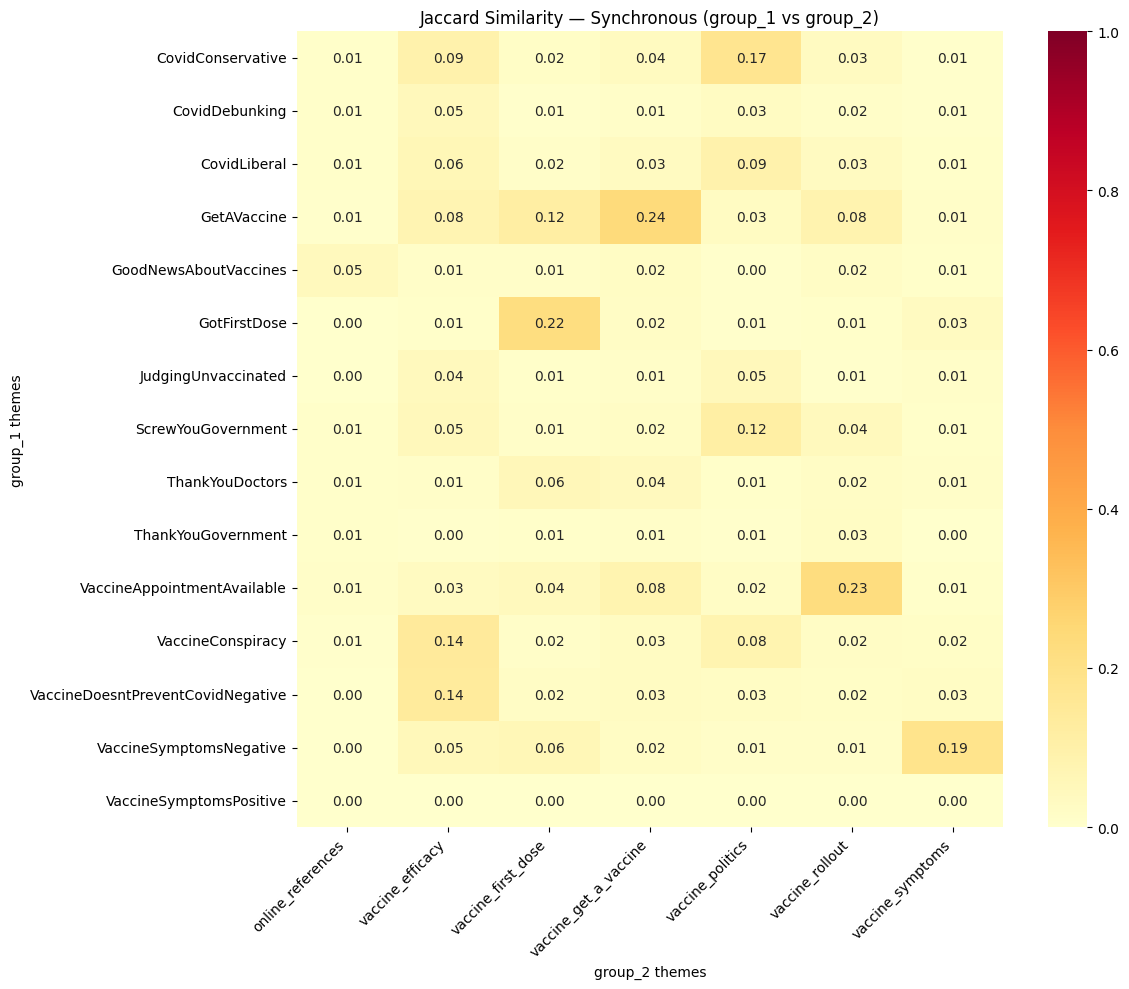}
        \caption{COVID tweets.}
        \label{fig:covid_laca_jaccard}
    \end{subfigure}
    \caption{Jaccard similarity heatmaps using the LLM-based approach.}
    \label{fig:jaccard_laca}
\end{figure*}

Heatmaps of the Jaccard similarity of themes from the synchronous experiments can be seen in figures \ref{fig:jaccard_fang}, \ref{fig:jaccard_pacheco}, and \ref{fig:jaccard_laca}. Each asynchronous experiment has three code-books so heatmaps between two code-books would not accurately reflect the Jaccard metrics reported in the paper.

\newpage

\section{Semi-Structured Interview}
\label{app:interview}

\subsection{Interviewing}
We administered interviews after annotation sessions. 
Asynchronous annotators were asked questions individually about their experience, whereas synchronous annotator groups were asked questions with their fellow annotators. 

\subsection{Script}

\begin{enumerate}
    \item Have you worked on annotation projects before? Did these annotation projects use qualitative coding strategies (ex: grounded theory)? How experienced are you as an annotator?  
    \item How was your experience on the COVID-19 vaccine annotation session we conducted on Sunday? Particularly, we are interested in your thoughts and feelings over the session.
    \item You annotated in a group, working together as a team. Did you find this setup to be beneficial? What were some of the limitations you faced, both individually and and as a group, when working synchronously?  
    \item  On a similar line, what would you consider to be the pros and cons if you were to annotate alone? 
\end{enumerate}

\textit{The last questions would be flipped based on if we are posing it to synchronous or asynchronous annotators. }

\section{Use of Pre-Existing Artifacts}

All pre-existing artifacts utilized in this study, including datasets, software libraries, models, and computational tools, are publicly available under open-source and open-access licenses. This academic work adheres to all intended use guidelines
and terms of service for the respective resources.

\end{document}